\documentclass{article}

\usepackage{iclr2027_conference,times}
\iclrfinalcopy

\usepackage[utf8]{inputenc} 
\usepackage[T1]{fontenc}    
\usepackage{hyperref}       
\usepackage{url}            
\usepackage{booktabs}       
\usepackage{amsfonts}       
\usepackage{nicefrac}       
\usepackage{microtype}      
\usepackage{xcolor}         
\usepackage{colortbl}       
\usepackage{amsmath}        
\usepackage{graphicx}       
\usepackage{algorithm}      
\usepackage{algorithmic}    
\usepackage{multirow}       
\usepackage{diagbox}        
\usepackage{subcaption}     
\usepackage{enumitem}       
\setlist[itemize]{itemsep=2pt, topsep=2pt, parsep=0pt}
\setlist[enumerate]{itemsep=2pt, topsep=2pt, parsep=0pt}

\definecolor{maroon}{cmyk}{0,0.87,0.68,0.32}
\definecolor{bamboo}{cmyk}{0.4,0,0.3,0}
\definecolor{apple}{cmyk}{0.41,0.4,0.76,0}
\definecolor{jialingshui}{cmyk}{0.47,0,0.49,0}
\definecolor{sea}{cmyk}{1,0.67,0.16,0.03}

\title{Forward-Forward Learning for Regression}

\author{%
Xinyang Liu$^{1}$, Xuanyu Liang$^{2}$, Shiqi Ding$^{3}$, Boyang Li$^{3}$, \\
\textbf{Zhiqiang Que$^{1}$, Jiayang Li$^{1}$, Guosheng Hu$^{1}$}\thanks{Corresponding author.} \\
{\normalfont $^{1}$University of Bristol \quad $^{2}$University College London \quad $^{3}$University of Cambridge} \\
{\normalfont\small\texttt{codex.lxy@gmail.com}, \texttt{xuanyu.liang.24@ucl.ac.uk}, \texttt{\{sd2089,bl574\}@cam.ac.uk},} \\
{\normalfont\small\texttt{\{z.que,jiayang.li,g.hu\}@bristol.ac.uk}}
}

\begin{document}

\maketitle
\lhead{Preprint}

\begin{abstract}
The Forward-Forward (FF) algorithm offers a computationally efficient and biologically plausible alternative to backpropagation (BP) by training neural networks through purely local, layer-wise optimization. However, FF is inherently designed for classification via contrastive positive-negative sample pairs, and extending it to regression poses fundamental challenges: \emph{continuous} target space lacks natural ``opposites'' for contrastive learning, and the standard goodness function carries no information about target magnitude or ordering. We propose \textbf{FFR} (Forward-Forward for Regression), to our knowledge, the first framework to extend FF to real-world regression and demonstrate competitive performance across diverse real-world datasets. FFR introduces three key innovations: (1) an ordinal competitive goodness function that replaces contrastive pairs with competitive learning between partitioned neuron groups under distance-aware ordinal supervision; (2) a stratified ladder architecture where shallow layers learn coarse ordinal discrimination and deeper layers refine into fine-grained regression, with multi-scale feature aggregation for inter-layer collaboration; and (3) hierarchical prediction with uncertainty estimation, where multi-scale predictors jointly provide robust predictions and a single-pass uncertainty score. Extensive experimental results show FFR recovers on average $98.5\%$ of BP's accuracy across six real-world regression benchmarks while reducing peak training memory to only $27\%$ of BP's at depth $8$ and $8\%$ at depth $32$, with per-iteration time around $72\%$ of BP's, and substantially outperforms all BP-free competitors.
\end{abstract}

\section{Introduction}
\label{sec:intro}
Backpropagation (BP) \citep{rumelhart1986learning} has served as the cornerstone of deep learning for decades. 
Despite its success, BP has well-known limitations. 
Computationally, it demands storing all intermediate activations (freezing activity) and prevents any layer from updating until the full forward-backward cycle completes (update locking) \citep{nokland2019training,bengio2006greedy,sun2025deeperforward}. 
Biologically, it requires symmetric weight transport between forward and backward passes and global error signals that propagate through the entire network, both of which are widely regarded as biologically implausible and difficult to implement efficiently on neuromorphic hardware \citep{lillicrap2016random,schuman2022opportunities,yi2023activity,zhu2025stochastic}. 
These constraints have motivated growing interest in biologically plausible alternatives that rely on local learning rules.

\begin{figure}[t]
  \centering
  \includegraphics[width=\linewidth]{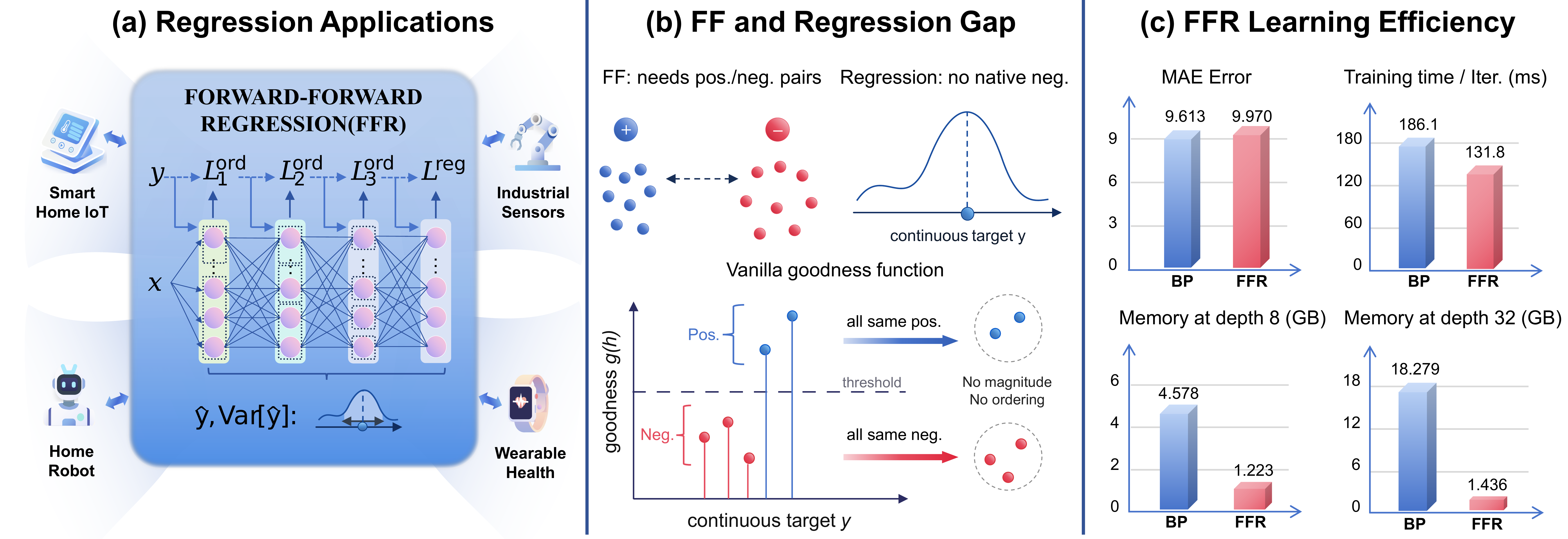}
  \vspace{-1.2em}
  \caption{Overview of FFR. (a) FFR framework and regression applications. (b) The gap between FF and regression. (c) Learning efficiency of FFR against BP on KonIQ-10k \citep{hosu2020koniq}: peak memory measured at depth $8$ and depth $32$, and per-iteration training time measured at depth $8$.}
  \label{fig:background}
  \vspace{-1.2em}
\end{figure}

Among the most prominent alternatives, \citet{hinton2022forward} proposed the Forward-Forward (FF) algorithm, which trains networks through purely local, layer-wise optimization with significantly lower memory footprint and reduced computational cost, making it attractive for resource-constrained hardware such as smart-home IoT sensors, industrial edge controllers, robotics, wearable monitors \citep{nahavandi2022application,chen2025self,wang2025empowering} (Figure~\ref{fig:background}(a)). FF replaces the backward pass with two forward passes, one on positive (real) data and one on negative (corrupted) data, training each layer independently by maximizing a local goodness function for positive samples while minimizing it for negatives. This eliminates intermediate-activation storage and enables fully layer-parallel weight updates, while satisfying biological plausibility.

However, FF is inherently designed for classification and has not been successfully applied to real-world regression, despite efficient regression learning being critical in domains with limited computation or restricted data sharing, such as private household sensor networks \citep{candanedo2017data}, proprietary industrial tool-wear monitoring \citep{de2023tool}, on-device indoor localization \citep{torres2014ujiindoorloc}, personal wearable monitoring \citep{pimentel2017toward}, and image quality assessment \citep{hosu2020koniq}. This gap stems from FF's contrastive mechanism \citep{chen2020simple}, which discriminates ``genuine'' input-label pairs (positives) from ``spurious'' ones (negatives): negatives are trivially constructed in classification by pairing inputs with random incorrect labels, but this cannot naturally extend to continuous targets. The difficulty is twofold (as shown in Figure~\ref{fig:background}(b)). \emph{Pairs construction}: in continuous target spaces, the boundary between correct and incorrect values is inherently ambiguous, with no natural way to define negatives. \emph{Goodness design}: the standard goodness measures squared activations for binary discrimination and carries no information about target magnitude or ordering, both essential for supervising real-valued predictions.

To our knowledge, only two concurrent \emph{preprints} attempt to bridge this gap. \citet{padmani2025function} cast regression as binary classification over probe points within a tolerance band, inheriting contrastive-pair overhead with limited accuracy: their method requires dozens of probe points per training sample and a sweep over hundreds of candidate targets at inference. \citet{guo2026local} replace FF's goodness with directional-derivative estimation for forward-only training on physical hardware, departing from the goodness-function paradigm and trading accuracy for implementability. 
Both works also restrict their evaluation to toy settings, namely low-dimensional synthetic function regression and MNIST classification recast as a regression target, on which accuracy falls short of BP. Neither establishes comparable accuracy on real-world regression. Whether FF's local learning can rival BP on real-world regression thus remains an open question.

To overcome these challenges, we propose \textbf{FFR} (Forward-Forward for Regression), a principled framework built on three key innovations: (1) an ordinal competitive goodness function that replaces contrastive pairs with competitive learning among partitioned neuron groups under distance-aware ordinal supervision; (2) a stratified ladder architecture where shallow layers learn coarse ordinal discrimination and deeper layers refine into fine-grained regression, with multi-scale feature aggregation for inter-layer collaboration;  (3) hierarchical prediction with uncertainty estimation that exploits the ladder's per-layer outputs to provide a single-pass uncertainty score. FFR retains FF's forward-only locality, yielding substantial memory and time savings over BP (Figure~\ref{fig:background}(c)). Our contributions are:
\begin{itemize}
    \item We propose FFR, to our knowledge, the first framework to extend FF to real-world regression and demonstrate competitive performance across diverse domains, including smart-home IoT, industrial sensing, indoor localization, wearable health, sentiment analysis, and image quality assessment, covering tabular, text, and image data.
    \item We propose three technical innovations: (1) an ordinal competitive goodness function that replaces contrastive pairs with competitive learning between grouped neurons under distance-aware ordinal supervision; (2) a stratified ladder architecture with depth-increasing neuron groups and multi-scale feature aggregation for inter-layer collaboration; (3) hierarchical prediction with uncertainty estimation that derives a single-pass uncertainty score from layer-wise predictor disagreement.
    \item FFR achieves on average $98.5\%$ of BP's performance on six real-world benchmarks while keeping peak training memory nearly constant with depth (about $27\%$ of BP's at depth $8$, $8\%$ at depth $32$) and average per-iteration time about $72\%$ BP's, substantially outperforming all existing BP-free competitors under matched network depth.
\end{itemize}

\section{Related Work}
\label{sec:related}

A rich family of BP-free methods has been proposed to address the limitations of backpropagation. Feedback alignment (FA) \citep{lillicrap2016random} and its direct variant (DFA) \citep{nokland2016direct} replace symmetric backward weights with random projections; target propagation \citep{bengio2014auto,lee2015difference}, equilibrium propagation \citep{scellier2017equilibrium}, and predictive coding \citep{whittington2017approximation} derive updates from layer-wise reconstructions, equilibrium states, or local prediction errors. PEPITA \citep{dellaferrera2022error} modulates the input with the output error in a second forward pass, while $\text{F}^3$ \citep{flugel2024f3} drives layer-wise updates through delayed and fixed random feedback paths. Perturbation-based methods \citep{jabri1992weight} estimate gradients by injecting noise. Most of these still require some form of global error signal, symmetric structure, or multi-phase computation that limits their locality.

Among fully local alternatives, the Forward-Forward (FF) algorithm \citep{hinton2022forward} replaces the backward pass with two forward passes on positive and negative data, training each layer with a local goodness function. Subsequent work extended FF to convolutional architectures via channel-wise competition \citep{papachristodoulou2024convolutional}, addressed inter-layer coordination \citep{lorberbom2024layer,ma2025ff}, introduced self-contrastive negatives \citep{chen2025self}, and tackled depth limitations \citep{sun2025deeperforward}, but all remain within the classification paradigm.

Two concurrent \emph{preprint} works attempt to extend FF to regression: \citet{padmani2025function} cast regression as binary classification over probe points within a tolerance band, retaining the contrastive paradigm with substantial probe-pair overhead, and FF-Zero \citep{guo2026local} replaces the FF goodness with directional-derivative estimation for forward-only training on analog hardware, departing from the goodness-function paradigm and trading accuracy for physical implementability.

To the best of our knowledge, this work is the first to extend FF to real-world regression and demonstrate competitive performance across diverse real-world datasets, whereas prior work has only been evaluated on toy settings and has not demonstrated comparable performance to conventional BP. Methodologically, our grouped-neuron competition shares the spirit of \citet{papachristodoulou2024convolutional} but repurposes it for regression with distance-aware ordinal labels, while our ladder architecture addresses inter-layer coordination \citep{lorberbom2024layer,ma2025ff} through multi-scale feature aggregation. In doing so, FFR retains FF's layer-local optimization: each hidden layer is trained with its own local objective, and the terminal regression head does not backpropagate gradients into the hidden layers.

\section{Methodology}
\label{sec:method}

\paragraph{Preliminaries.}
\label{sec:prelim}
The FF algorithm \citep{hinton2022forward} was originally proposed for classification, training each layer to discriminate between correctly and incorrectly labeled inputs. Given a feedforward network with $L$ hidden layers, an input $\mathbf{x} \in \mathbb{R}^d$ and its class label $y \in \{1, \dots, C\}$, FF overwrites the first $C$ dimensions of $\mathbf{x}$ with a one-hot encoding of $y$ to produce $\mathbf{x}^+ = \texttt{embed}(\mathbf{x}, y)$. At layer $\ell$, the activation is $\mathbf{h}_\ell = \sigma( W_\ell \, \bar{\mathbf{h}}_{\ell-1} + \mathbf{b}_\ell )$ with $\bar{\mathbf{h}}_{\ell-1} = \mathbf{h}_{\ell-1} / \|\mathbf{h}_{\ell-1}\|$, and the \emph{goodness} is $g_\ell = \|\mathbf{h}_\ell\|^2$. A negative sample $\mathbf{x}^- = \texttt{embed}(\mathbf{x}, \tilde{y})$ uses a random incorrect label $\tilde{y} \neq y$, and each layer minimizes:
\begin{equation}
\label{eq:ff_loss}
    \mathcal{L}_\ell = \log\!\left(1 + e^{-(g_\ell^+ - \theta)}\right) + \log\!\left(1 + e^{(g_\ell^- - \theta)}\right),
\end{equation}
where $\theta$ is a threshold. At inference time, FF evaluates all $C$ label embeddings and selects $\hat{y} = \arg\max_{c} \sum_{\ell} g_\ell^{(c)}$. This mechanism is fundamentally discrete: it requires enumerating all candidate labels, and the goodness $g_\ell$ carries no information about magnitude or ordering, which are important for regression. For a continuous target $y \in \mathbb{R}$, enumeration is impossible, and constructing a meaningful negative is ill-defined, as $y + 0.1$ and $y + 100$ cannot both be treated as equally incorrect.

To lift FF beyond its classification roots, we propose \textbf{FFR} (Forward-Forward Learning for Regression), the first framework to effectively extend FF to real-world regression. FFR addresses the two fundamental difficulties through three components: (1) an ordinal competitive goodness function (Section~\ref{sec:goodness}); (2) a stratified ladder architecture (Section~\ref{sec:ladder}); and (3) a hierarchical prediction with uncertainty estimation (Section~\ref{sec:hp}). We detail each in turn below.

\begin{figure}[t]
\centering
\includegraphics[width=\linewidth]{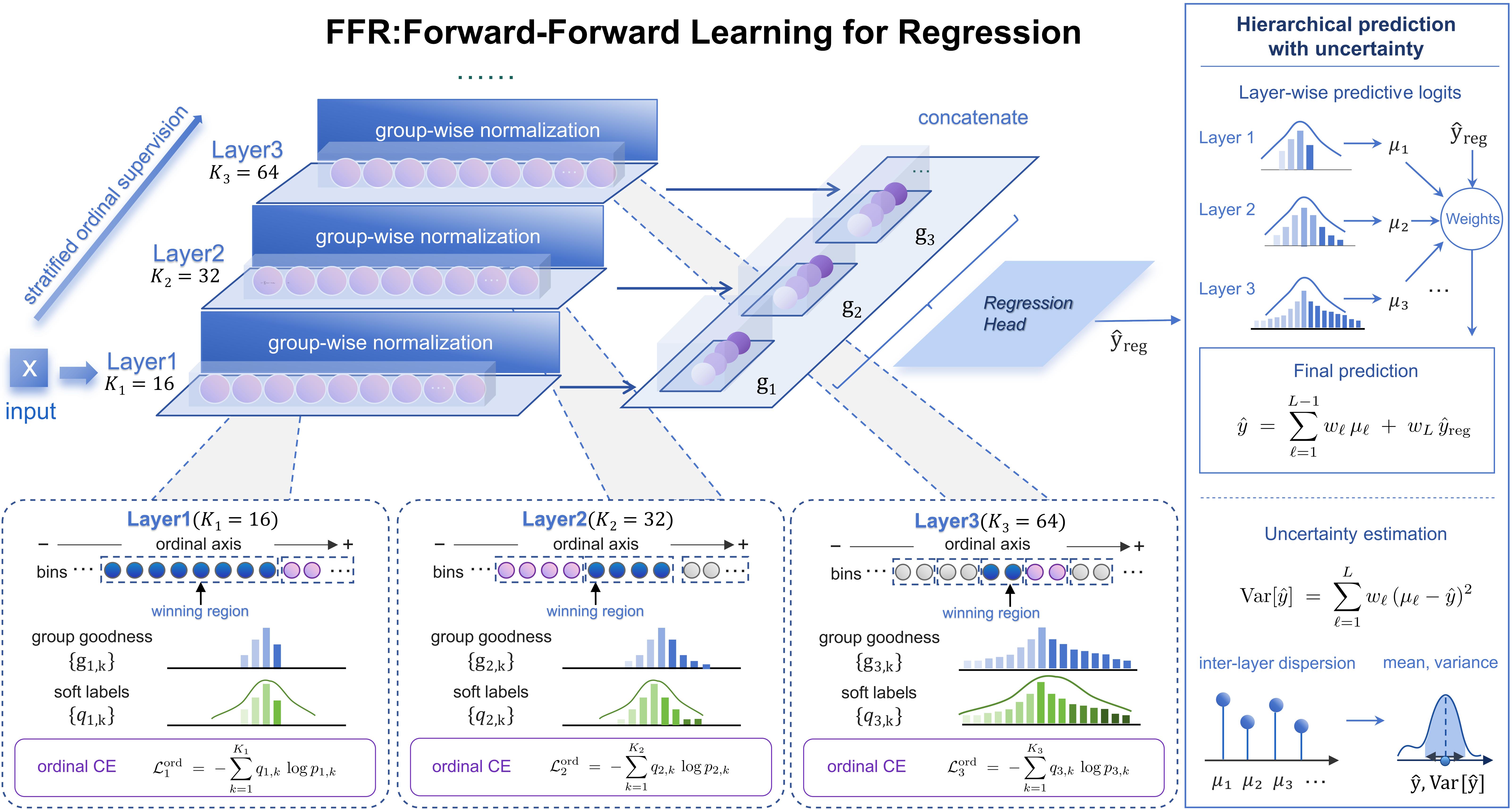}
\vspace{-1.2em}
\caption{FFR framework. Stratified ladder architecture is trained with ordinal competitive goodness function, then ensembled into $\hat{y}$ with a single-pass uncertainty score.}
\vspace{-1em}
\label{fig:framework}
\end{figure}

\subsection{Ordinal Competitive Goodness Function}
\label{sec:goodness}
A direct MSE goodness surrogate on each layer is the most intuitive adaptation of FF to regression: for layer $\ell$ with $n_\ell$ hidden units and activations $\{h_{\ell,j}\}_{j=1}^{n_\ell}$, we average the activations into a scalar estimate $\hat{y}_\ell = \tfrac{1}{n_\ell}\sum_{j=1}^{n_\ell} h_{\ell,j}$ and minimize $\mathcal{L}_\ell = (\hat{y}_\ell - y)^2$, yielding a naive FF-MSE baseline without probe pairs or inference sweeps.
Yet this formulation breaks FF's core competitive mechanism. 
Standard FF trains neurons to raise the goodness of correct representations and suppress incorrect ones, whereas MSE reduces each layer to a greedy point estimator by matching its activation average to the target. 
Competitive pressure vanishes, activations collapse to redundant features, and shallow layers overfit and pass collapsed representations downstream, leaving no room for further refinement. 
We term this failure representation collapse under greedy regression and empirically confirm its severity in Appendix~\ref{app:results:dynamics}. An effective FF regression objective should instead preserve local competition and encode the ordinal structure of continuous targets, rather than regress a scalar per layer.

Motivated by these requirements, we propose the \emph{ordinal competitive goodness function} (as shown in the bottom of Figure~\ref{fig:framework}), which replaces FF's positive/negative-pair contrast with an intra-layer competition over a discretized target space: each hidden layer becomes an ordinal classifier where disjoint neuron groups compete to represent different regions of the continuous target range. Concretely, given a target range $[y_{\min}, y_{\max}]$, we partition it at layer $\ell$ into $K_\ell$ equal-width ordered bins of width $\Delta_\ell = (y_{\max}-y_{\min})/K_\ell$, with midpoints $m_{\ell,k} = y_{\min} + (k-\tfrac{1}{2})\Delta_\ell$ for $k=1,\dots,K_\ell$; each target $y$ thus lies in a unique bin $k^\star(y)$ centred at $m_{\ell,k^\star(y)}$.

Correspondingly, we partition the $n_\ell$ hidden units (or channels for CNN) of layer $\ell$ into disjoint \emph{competitive groups} $\{\mathcal{G}_{\ell,1},\dots,\mathcal{G}_{\ell,K_\ell}\}$, each with goodness given by its mean squared activation,
\begin{equation}
\label{eq:group_goodness}
    g_{\ell,k} \;=\; \frac{1}{|\mathcal{G}_{\ell,k}|} \sum_{j \in \mathcal{G}_{\ell,k}} h_{\ell,j}^2,
\end{equation}
which is then normalized into a distribution over bins, $p_{\ell,k}=\exp(g_{\ell,k}) /\sum_{k'=1}^{K_\ell}\exp(g_{\ell,k'})$. A group ``wins'' when its neurons fire most strongly among all groups; training a layer thus reduces to shaping which group wins as a function of the target, without requiring any negative sample construction. Since all groups share the same input, they implicitly compete for representational capacity.

However, supervising this competition with a hard one-hot label on $k^\star(y)$ would ignore the ordinal structure of the target space: two predictions off by one bin and off by twenty bins would be penalized identically. Instead, we construct a distance-aware soft label that places a Gaussian bump around $y$ in target units and then projects it onto the bin midpoints,
\begin{equation}
\label{eq:soft_label}
    q_{\ell,k} \;=\; \frac{\exp\!\big(-(y - m_{\ell,k})^2 / (2\sigma_\ell^2)\big)}{\sum_{k'=1}^{K_\ell} \exp\!\big(-(y - m_{\ell,k'})^2 / (2\sigma_\ell^2)\big)},
\end{equation}
with the bandwidth tied to the bin width, $\sigma_\ell = \Delta_\ell$, so that nearby bins always receive non-trivial mass while distant bins are effectively suppressed. The goodness function per layer is the cross-entropy between this ordinal soft label $q_{\ell,k}$ and the normalized goodness distribution $p_{\ell,k}$ over the $K_\ell$ groups,
\begin{equation}
\label{eq:ordinal_ce}
    \mathcal{L}_\ell^{\mathrm{ord}} \;=\; -\sum_{k=1}^{K_\ell} q_{\ell,k}\,\log p_{\ell,k},
\end{equation}
which simultaneously raises the goodness of the target bin and its neighbours while suppressing distant bins in proportion to their distance from $y$. This replaces standard FF's positive/negative-pair contrast with a local intra-layer competition supervised by the continuous ordering of targets.

Without loss of generality, for a $D$-dimensional target $\mathbf{y}\in\mathbb{R}^D$, the available groups are partitioned across output dimensions to give midpoints $m_{\ell,k}^{(d)}$ and bin indices $k^\star_d(\mathbf{y})$, and all subsequent designs are replicated per dimension. Layer parameters are shared across all $D$ targets, and the local objective is averaged over the $D$ per-dimension terms that jointly update this shared representation (see Appendix~\ref{app:multitarget} for details).

\subsection{Stratified Ladder Architecture}
\label{sec:ladder}
The top-left of Figure~\ref{fig:framework} depicts the stratified ladder architecture, which progresses from coarse to fine ordinal partitions across depth and aggregates multi-scale goodness into a terminal regression head.
A fixed group number $K$ shared across all layers would force every layer to operate at the same ordinal granularity and therefore to learn near-identical features, which again invites collapse. We instead stratify the discretization across depth by doubling the number of competitive groups every layer,
\begin{equation}
\label{eq:K_schedule}
    K_\ell \;=\; 2^{d_0 + \ell - 1}, \qquad \ell = 1,\dots,L-1,
\end{equation}
with a base exponent $d_0$ ($d_0 = 4$ in our experiments). Shallow layers thus carve the target range into few, wide bins that are easy to separate from the raw input, while deeper layers inherit those coarse anchor points and refine them into progressively finer partitions. Combined with Eq.~\eqref{eq:ordinal_ce}, this stratification lets each intermediate layer represent a distribution over ordinal target bins, rather than a point estimate, that deeper layers can still reshape into finer-grained predictions.

To prevent one group from ``leaking'' into another by simply inflating its activations, normalization is applied \emph{within each group} rather than across the whole layer. Concretely, if $\mathbf{h}_\ell = (\mathbf{h}_{\ell,1},\dots,\mathbf{h}_{\ell,K_\ell})$ denotes the partitioned activations, we apply a group-wise normalization:
\begin{equation}
\label{eq:group_norm}
    \tilde{h}_{\ell,j} \;=\; \gamma_{\ell,j}\,\frac{h_{\ell,j} - \mu_{\ell,k}}{\sqrt{\sigma^2_{\ell,k} + \varepsilon}} + \beta_{\ell,j}, \qquad j\in\mathcal{G}_{\ell,k},
\end{equation}
where $\mu_{\ell,k},\sigma^2_{\ell,k}$ are the per-sample mean and variance computed only inside group $\mathcal{G}_{\ell,k}$, and $\gamma_{\ell,j},\beta_{\ell,j}$ are learnable per-unit affine parameters. The goodness in Eq.~\eqref{eq:group_goodness} is computed on these normalized activations. Because each group's statistics are computed independently, the activation range of every group becomes comparable when passed to the next layer, forcing the subsequent layer to re-encode the target information \emph{within} its own feature subspace. Combined with the ordinal soft label, this group-wise normalization stabilizes training and helps prevent overfitting.

Finally, to coordinate layers without breaking FF's local-learning property, we use a ladder architecture in which the grouped goodness (low memory compared to activations) of all intermediate layers are concatenated and fed to a terminal regression layer that predicts the continuous target under MSE:
\begin{equation}
\label{eq:ladder_reg}
    \hat{y}_{\mathrm{reg}} \;=\; W_{\mathrm{reg}}\,\big[\,\mathbf{g}_1;\; \mathbf{g}_2;\; \dots;\; \mathbf{g}_{L-1}\,\big] + b_{\mathrm{reg}}, \qquad \mathcal{L}^{\mathrm{reg}} \;=\; (\hat{y}_{\mathrm{reg}} - y)^2.
\end{equation}
The terminal layer is the only place where gradients mix features across layers, and those gradients are confined to the single linear map $W_{\mathrm{reg}}$: no gradient flows back through the intermediate layers, which remain trained purely by their own $\mathcal{L}^{\mathrm{ord}}_\ell$. The stratified ladder architecture therefore delivers inter-layer collaboration while preserving the update-locality and backpropagation-free nature of FF.

\subsection{Hierarchical Prediction with Uncertainty Estimation}
\label{sec:hp}
As shown in the right of Figure~\ref{fig:framework}, a notable advantage of the stratified ladder architecture is that each intermediate layer already carries a complete predictive distribution over the target. Because layer $\ell$ trained with $\mathcal{L}^{\mathrm{ord}}_\ell$ outputs a softmax $p_{\ell,k}$ over the $K_\ell$ bin midpoints $m_{\ell,k}$, we can extract a continuous intermediate prediction directly from its own activations,
\begin{equation}
\label{eq:layer_moment}
    \mu_\ell \;=\; \sum_{k=1}^{K_\ell} p_{\ell,k}\, m_{\ell,k},
\end{equation}
so that shallower layers provide robust predictions while deeper layers provide sharper, more localized estimates. Together with the terminal regression output $\hat{y}_{\mathrm{reg}}$ from Eq.~\eqref{eq:ladder_reg}, this gives $L$ jointly available predictors at multiple ordinal scales, all obtained in a single forward pass.

We combine these predictors into the final, robust estimate $\hat{y}$ via a weighted hierarchical ensemble,
\begin{equation}
\label{eq:ladder_pred}
    \hat{y} \;=\; \sum_{\ell=1}^{L-1} w_\ell\, \mu_\ell \;+\; w_L\, \hat{y}_{\mathrm{reg}}, \qquad \sum_{\ell=1}^{L} w_\ell = 1, \quad w_\ell \geq 0,
\end{equation}
Here, $\hat{y}_{\mathrm{reg}}$ is the continuous terminal prediction and is not limited by the bin resolution, while $\hat{y}$ combines it with the coarser layer-wise predictors. The weights $w_\ell$ can be set to uniform values or calibrated on a held-out split. Ensembling coarse predictors from early layers with late layers reduces the impact of any single layer's idiosyncratic errors and smooths out overfitting on small datasets. More importantly, disagreement among these layer-wise predictors provides a single-pass uncertainty score. We quantify this disagreement by the weighted variance across predictors,
\begin{equation}
\label{eq:uncertainty}
    U(\mathbf{x}) \;=\; \sum_{\ell=1}^{L} w_\ell\,(\mu_\ell - \hat{y})^2,
\end{equation}
where we set $\mu_L = \hat{y}_{\mathrm{reg}}$ for the terminal head. This disagreement-based score $U(\mathbf{x})$ is used as an estimate of epistemic uncertainty, reflecting variation in the learned predictor rather than irreducible observation noise~\citep{kendall2017uncertainties}. Predictor disagreement is widely used for uncertainty estimation in approximate Bayesian and ensemble-based methods~\citep{gal2016dropout,lakshminarayanan2017simple,jung2023a}. In FFR, the disagreement arises among predictors derived from different depths and ordinal resolutions within a single deterministic forward pass. Under squared loss, the same weighted disagreement appears as the ambiguity term in the classical ensemble decomposition~\citep{krogh1994neural}: the ensemble error equals the weighted average member error minus this term. A derivation is provided in Appendix~\ref{app:uncertainty}, with quantitative evaluation in Appendix~\ref{app:uncertainty_eval}. For multi-target regression, Eqs.~\eqref{eq:layer_moment}--\eqref{eq:uncertainty} are evaluated per output dimension with a $D$-dimensional terminal head (Appendix~\ref{app:multitarget}).

\section{Experiments}
\label{sec:exp}

\subsection{Experimental settings}
We evaluate FFR on four synthetic function regression tasks (\emph{Sin-Cos} and \emph{Exp-Trig-Poly} from \citet{guo2026local}, and their multi-target variants MT-A and MT-B), plus six real-world datasets (Appliances Energy \citep{candanedo2017data}, Machine Tool Wear \citep{de2023tool}, UJIIndoorLoc \citep{torres2014ujiindoorloc}, BIDMC \citep{pimentel2017toward}, CMU-MOSEI \citep{zadeh2018multimodal}, and KonIQ-10k \citep{hosu2020koniq}) spanning smart home IoT, industrial manufacturing, indoor localization, wearable health, sentiment analysis, and image quality assessment; dataset details are provided in Appendix~\ref{app:datasets}.
We compare FFR against two BP references (BP-UR, the same FFR architecture trained globally with backpropagation, and BP-EX, BP with an extra per-layer ordinal loss), two intuitive FF baselines with grouped neurons supervised per layer by an MSE loss (FF-MSE) or a one-hot classification loss (FF-CLF), three existing FF methods (FF-CAR \citep{padmani2025function}, Trifecta \citep{dooms2023trifecta}, and FF-Zero \citep{guo2026local}), and two non-FF BP-free methods (PEPITA \citep{dellaferrera2022error} and $\text{F}^3$ \citep{flugel2024f3}); all classification-style baselines discretize the target into 64 bins and recover it from the logits. The synthetic and four tabular datasets share a 4-layer fully connected backbone of width $256$, KonIQ-10k uses an 8-layer CNN backbone, and CMU-MOSEI uses a 3-block GRU backbone of width $256$ for recurrent modeling. FFR uses group numbers $K_\ell \in \{16,32,64\}$ across layers and uniform hierarchical-prediction weights $w_\ell = 1/L$. For a fair comparison, all methods on each dataset share the same optimization setting and data split protocol; per-dataset learning rates, batch sizes, epochs, and other implementation details are provided in Appendices~\ref{app:datasets}--\ref{app:details}.

\begin{table}[t]
\vspace{-1em}
  \caption{Test RMSE ($\downarrow$) and MAE ($\downarrow$) on synthetic function regression benchmarks. \textbf{Bold}: best BP-free; \underline{underline}: second best.}
  \label{tab:synthetic}
  \vspace{-0.5em}
  \centering
  \setlength{\tabcolsep}{8.5pt}%
  \resizebox{\textwidth}{!}{%
    \begin{tabular}{l cc cc cc cc cc}
    \toprule\toprule
    \multirow{2}{*}{\diagbox[width=6.2em]{\textbf{Meth.}}{\textbf{Data}}} & \multicolumn{2}{c}{Sin-Cos} & \multicolumn{2}{c}{Exp-Trig-Poly} & \multicolumn{2}{c}{MT-A} & \multicolumn{2}{c}{MT-B} & \multicolumn{2}{c}{Average} \\
    \cmidrule(lr){2-3} \cmidrule(lr){4-5} \cmidrule(lr){6-7} \cmidrule(lr){8-9} \cmidrule(lr){10-11}
    & RMSE & MAE & RMSE & MAE & RMSE & MAE & RMSE & MAE & RMSE & MAE \\
    \midrule
    \rowcolor{jialingshui!10} BP-UR                                       & 0.005             & 0.004             & 0.007             & 0.005             & 0.008             & 0.005             & 0.011             & 0.008             & 0.008             & 0.006 \\
    \rowcolor{jialingshui!10} BP-EX                                       & 0.013             & 0.010             & 0.014             & 0.011             & 0.012             & 0.008             & 0.019             & 0.015             & 0.015             & 0.011 \\
    \midrule
    \rowcolor{maroon!5} $\text{F}^3$                & 0.096             & 0.074             & 0.078             & 0.059             & 0.067             & 0.047             & \underline{0.091} & \underline{0.071} & 0.083             & 0.063 \\
    \rowcolor{maroon!5} PEPITA            & 0.490             & 0.429             & 0.106             & 0.073             & 0.112             & 0.070             & 0.206             & 0.162             & 0.229             & 0.184 \\
    \rowcolor{maroon!5} Trifecta               & 0.026             & 0.021             & 0.035             & 0.025             & \underline{0.048} & \underline{0.031} & 0.095             & 0.078             & \underline{0.051} & \underline{0.039} \\
    \rowcolor{maroon!5} FF-CAR               & 0.070             & 0.054             & 0.057             & 0.042             & 0.267             & 0.171             & 0.345             & 0.259             & 0.185             & 0.131 \\
    \rowcolor{maroon!5} FF-Zero                     & 0.137             & 0.109             & 0.140             & 0.101             & 0.151             & 0.111             & 0.231             & 0.186             & 0.165             & 0.127 \\
    \rowcolor{maroon!5} FF-MSE                                            & 0.196             & 0.155             & 0.152             & 0.114             & 0.172             & 0.115             & 0.253             & 0.202             & 0.193             & 0.147 \\
    \rowcolor{maroon!5} FF-CLF                                            & \underline{0.010} & \underline{0.007} & \underline{0.027} & \underline{0.019} & 0.108             & 0.071             & 0.365             & 0.306             & 0.128             & 0.101 \\
    \midrule
    \rowcolor{sea!10} FFR (ours)                                          & \textbf{0.005}    & \textbf{0.004}    & \textbf{0.009}    & \textbf{0.006}    & \textbf{0.015}    & \textbf{0.010}    & \textbf{0.021}    & \textbf{0.017}    & \textbf{0.013}    & \textbf{0.009} \\
    \bottomrule\bottomrule
  \end{tabular}%
  }
\vspace{-1.2em}
\end{table}

\subsection{Main results}
As shown in Tables~\ref{tab:synthetic} and~\ref{tab:realworld}, FFR achieves the best performance among all BP-free methods on every benchmark, while remaining within a small margin of standard BP. On the four synthetic tasks, including the multi-target MT-A and MT-B, FFR's RMSE stays at or below $0.021$, averaging $0.013$ against BP-UR's $0.008$. On the six real-world datasets, FFR recovers about $98.5\%$ of BP-UR's performance, computed as the mean per-cell BP-UR-to-FFR error ratio over the twelve RMSE and MAE entries. It even surpasses BP-UR on UJIIndoorLoc, BIDMC, and Appliances (RMSE), indicating that hierarchical prediction adds complementary signal beyond end-to-end optimization.

The intuitive FF baselines confirm our analysis: FF-MSE underperforms FFR on every benchmark, with synthetic RMSE more than an order of magnitude worse, consistent with the representation collapse in Appendix~\ref{app:results:dynamics}; FF-CLF is competitive on the single-target tasks but its error grows by more than an order of magnitude on MT-A and MT-B, where each target must be discriminated across independent fine-grained bins. FFR avoids both failures via its stratified ladder and a continuous regression head unbound by bin resolution. The remaining baselines are each limited by their core design: FF-CAR \citep{padmani2025function} is not formulated for image inputs, and its in-/out-tol probe sweep makes training and inference prohibitively expensive; FF-Zero \citep{guo2026local} trades accuracy for hardware feasibility via a directional-derivative goodness; Trifecta \citep{dooms2023trifecta} is bound by output bin resolution; $\text{F}^3$'s \citep{flugel2024f3} fixed random feedback paths uniformly trail the FF family; and PEPITA's \citep{dellaferrera2022error} error-modulated forward signal does not transfer across modalities (BIDMC vs.\ Appliances).

\begin{table}[h]
\vspace{0.2em}
  \caption{Test RMSE / MAE ($\downarrow$) on real-world regression benchmarks. \textbf{Bold}: best BP-free; \underline{underline}: second best; --: not applicable (FF-CAR does not support image inputs).}
  \label{tab:realworld}
  \vspace{-0.5em}
  \centering
  \resizebox{\textwidth}{!}{%
  {\renewcommand{\arraystretch}{1.12}%
  \begin{tabular}{l!{\vrule width 0.8pt}c c c c c c}
    \toprule\toprule
    \textbf{Method} & Appliances & Tool Wear & UJIIndoorLoc & BIDMC & CMU-MOSEI & KonIQ-10k \\
    \midrule
    \rowcolor{jialingshui!10} BP-UR & 78.457 / 37.366 & 12.308 / 7.259 & 0.018 / 0.012 & 3.852 / 2.671 & 0.892 / 0.679 & 12.357 / 9.613 \\
    \rowcolor{jialingshui!10} BP-EX & 77.457 / 35.987 & 12.237 / 7.056 & 0.020 / 0.014 & 4.291 / 2.896 & 0.925 / 0.718 & 12.299 / 9.535 \\
    \midrule
    \rowcolor{maroon!5} $\text{F}^3$ & 93.832 / 54.004 & 58.834 / 37.188 & 0.138 / 0.107 & 4.088 / 3.018 & 0.959 / 0.744 & 15.158 / 11.097 \\
    \rowcolor{maroon!5} PEPITA & 94.080 / 49.335 & 33.100 / 21.230 & 0.022 / 0.014 & 4.818 / 3.105 & 1.005 / 0.771 & 17.044 / 13.088 \\
    \rowcolor{maroon!5} Trifecta & 84.660 / 52.242 & \underline{27.095} / \underline{15.864} & 0.048 / 0.039 & 4.400 / \underline{2.799} & 1.021 / 0.784 & 14.271 / 10.523 \\
    \rowcolor{maroon!5} FF-CAR & 127.587 / 71.092 & 115.072 / 65.310 & 0.604 / 0.532 & 5.291 / 2.863 & 1.356 / 0.999 & -- / -- \\
    \rowcolor{maroon!5} FF-Zero & 98.839 / 60.540 & 65.413 / 45.106 & 0.064 / 0.049 & \underline{3.911} / 2.864 & 1.090 / 0.827 & 13.764 / 10.766 \\
    \rowcolor{maroon!5} FF-MSE & 85.256 / \underline{44.963} & 39.815 / 18.526 & \underline{0.021} / \underline{0.014} & 4.767 / 3.231 & \underline{0.940} / \underline{0.737} & 14.134 / 11.183 \\
    \rowcolor{maroon!5} FF-CLF & \underline{84.608} / 45.117 & 39.376 / 22.808 & 0.027 / 0.015 & 4.993 / 3.227 & 1.088 / 0.829 & \underline{13.404} / \underline{10.351} \\
    \midrule
    \rowcolor{sea!10} FFR (ours) & \textbf{76.777} / \textbf{41.410} & \textbf{13.547} / \textbf{8.388} & \textbf{0.017} / \textbf{0.011} & \textbf{3.590} / \textbf{2.652} & \textbf{0.914} / \textbf{0.691} & \textbf{12.731} / \textbf{9.970} \\
    \bottomrule\bottomrule
  \end{tabular}%
  }%
  }
\vspace{0.3em}
\end{table}

\subsection{Ablations and analysis}
\label{sec:ablation}
This section provides ablation studies and more analyses of FFR. We first isolate the contribution of each proposed component (ordinal soft label, stratified group schedule, ladder aggregation, hierarchical prediction), and probe the model's behavior under varied $K_\ell$ schedules and hierarchical-prediction weights; both are evaluated on the four real-world tabular datasets and CMU-MOSEI with test RMSE and MAE reported. We then compare FFR's training memory and per-iteration time against BP as model depth grows, and visualize the single-pass uncertainty score derived from the ladder hierarchy.

\paragraph{Component ablation.}
Table~\ref{tab:ab_component} isolates the contribution of each component.
For the \emph{ordinal competitive goodness} (OCG), the FF-MSE baseline (Tables~\ref{tab:synthetic}--\ref{tab:realworld}) already isolates the competitive design, so we only replace the ordinal soft label with one-hot supervision (\emph{w/o soft label}).
For the \emph{stratified ladder architecture}, we ablate stratification by fixing $K_\ell$ across all layers to either the smallest (\emph{w/o strat.~$(16^3)$}) or largest (\emph{w/o strat.~$(64^3)$}) group count used by Full FFR, and separately ablate ladder aggregation (\emph{w/o ensemble}: the terminal head sees only the last layer).
For \emph{hierarchical prediction} (\emph{w/o HP}), we drop the weighted multi-layer ensemble and use the terminal head alone.
Full FFR achieves the best result in every column: removing the soft label, stratification, ladder aggregation, or hierarchical prediction each degrades performance, indicating complementary roles across both feed-forward and recurrent backbones. Layer-wise visualizations on KonIQ-10k further show that the blocks learn progressively different spatial responses and ordinal distributions, complementing the CKA evidence of cross-layer differentiation (Appendix~\ref{app:results:dynamics}, Figures~\ref{fig:koniq_spatial}--\ref{fig:koniq_bins}).

\begin{table}[t]
\vspace{0em}
  \caption{Component ablation on real-world datasets. Test RMSE ($\downarrow$) and MAE ($\downarrow$). \textbf{Bold}: best per column; \underline{underline}: second-best.}
  \label{tab:ab_component}
  \centering
  \resizebox{\textwidth}{!}{%
    \begin{tabular}{l cc cc cc cc cc}
    \toprule\toprule
        \multirow{2}{*}{\diagbox[width=8em]{\textbf{Variants}}{\textbf{Data}}}\hspace{0.6em} & \multicolumn{2}{c}{Appliances} & \multicolumn{2}{c}{Tool Wear} & \multicolumn{2}{c}{UJIIndoorLoc} & \multicolumn{2}{c}{BIDMC} & \multicolumn{2}{c}{CMU-MOSEI} \\
    \cmidrule(lr){2-3} \cmidrule(lr){4-5} \cmidrule(lr){6-7} \cmidrule(lr){8-9} \cmidrule(lr){10-11}
    & RMSE & MAE & RMSE & MAE & RMSE & MAE & RMSE & MAE & RMSE & MAE \\
    \midrule
    \rowcolor{maroon!5} w/o soft label                                  & 81.787 & 44.862 & 31.380 & 20.120 & 0.048 & 0.033 & 4.526 & 3.693 & 0.9222 & 0.7040 \\
    \rowcolor{maroon!5} w/o strat.~$(16^3)$                             & 79.399 & 45.853 & 15.454 & 10.274 & 0.041 & 0.030 & 4.727 & 3.554 & 0.9168 & \underline{0.6981} \\
    \rowcolor{maroon!5} w/o strat.~$(64^3)$                             & 82.012 & 45.073 & 29.989 & 17.967 & 0.038 & 0.026 & 4.406 & 3.359 & \underline{0.9139} & 0.7003 \\
    \rowcolor{maroon!5} w/o ensemble                                    & \underline{79.179} & 44.343 & 26.032 & 14.848 & 0.037 & 0.027 & 4.502 & 3.547 & 0.9332 & 0.7041 \\
    \rowcolor{maroon!5} w/o HP                                          & 79.186 & \underline{42.630} & \underline{15.048} & \underline{9.791} & \underline{0.033} & \underline{0.023} & \underline{4.269} & \underline{3.242} & 0.9239 & 0.7009 \\
    \midrule
    \rowcolor{sea!10} Full FFR                                          & \textbf{76.777} & \textbf{41.410} & \textbf{13.547} & \textbf{8.388} & \textbf{0.017} & \textbf{0.011} & \textbf{3.590} & \textbf{2.652} & \textbf{0.9138} & \textbf{0.6947} \\
    \bottomrule\bottomrule
  \end{tabular}%
  }
\vspace{-1em}
\end{table}

\paragraph{Hyperparameter sensitivity.}
Table~\ref{tab:ab_hparam} reports sensitivity along two axes: (i) the group schedule $K_\ell$ and (ii) the hierarchical prediction weighting scheme.
Performance is broadly stable across $K_\ell$ schedules, with $\{16,32,64\}$ providing the strongest overall balance and the best RMSE and MAE on CMU-MOSEI; the finer $\{32,64,128\}$ schedule degrades more noticeably on Tool Wear. For HP weighting we compare three configurations, config.~1 uniform $\{0.25,0.25,0.25,0.25\}$, config.~2 last-layer-dominant $\{0.1,0.1,0.1,0.7\}$, and config.~3 first-layer-dominant $\{0.7,0.1,0.1,0.1\}$. Uniform weighting remains the strongest overall choice, while the alternatives produce only small changes on CMU-MOSEI and mixed dataset-specific effects elsewhere.

\begin{table}[h]
  \vspace{-0.8em}
  \caption{Hyperparameter sensitivity on real-world datasets. Test RMSE ($\downarrow$) and MAE ($\downarrow$). Default FFR configuration is highlighted. \textbf{Bold}: best per column within each block; \underline{underline}: second-best.}
  \label{tab:ab_hparam}
  \centering
  \resizebox{\textwidth}{!}{%
    \begin{tabular}{l cc cc cc cc cc}
    \toprule\toprule
    \multirow{2}{*}{\diagbox[width=8em]{\textbf{Variants}}{\textbf{Data}}} & \multicolumn{2}{c}{Appliances} & \multicolumn{2}{c}{Tool Wear} & \multicolumn{2}{c}{UJIIndoorLoc} & \multicolumn{2}{c}{BIDMC} & \multicolumn{2}{c}{CMU-MOSEI} \\
    \cmidrule(lr){2-3} \cmidrule(lr){4-5} \cmidrule(lr){6-7} \cmidrule(lr){8-9} \cmidrule(lr){10-11}
    & RMSE & MAE & RMSE & MAE & RMSE & MAE & RMSE & MAE & RMSE & MAE \\
    \midrule
    \multicolumn{11}{l}{\emph{$K_\ell$ per layer}} \\
    \rowcolor{maroon!5}   \hspace{1em} $\{4,8,16\}$              & 77.246 & 42.283 & 14.761 & 8.933 & 0.020 & 0.014 & 4.888 & 3.260 & 0.9172 & 0.6970 \\
    \rowcolor{maroon!5}   \hspace{1em} $\{8,16,32\}$             & \textbf{76.201} & \textbf{40.023} & \textbf{12.840} & \underline{8.455} & \underline{0.019} & \underline{0.013} & \underline{4.502} & \underline{2.956} & 0.9294 & 0.7132 \\
    \rowcolor{sea!10}     \hspace{1em} $\{16,32,64\}$            & \underline{76.777} & \underline{41.410} & \underline{13.547} & \textbf{8.388} & \textbf{0.017} & \textbf{0.011} & \textbf{3.590} & \textbf{2.652} & \textbf{0.9138} & \textbf{0.6947} \\
    \rowcolor{maroon!5}   \hspace{1em} $\{32,64,128\}$           & 82.036 & 44.831 & 16.901 & 11.756 & 0.021 & 0.015 & 4.620 & 3.087 & \underline{0.9169} & \underline{0.6955} \\
    \midrule
    \multicolumn{11}{l}{\emph{HP weights}} \\
    \rowcolor{sea!10}     \hspace{1em} Config.~1   & \underline{76.777} & \underline{41.410} & 13.547 & 8.388 & \textbf{0.017} & \textbf{0.011} & \textbf{3.590} & \textbf{2.652} & \textbf{0.9138} & \textbf{0.6947} \\
    \rowcolor{maroon!5}   \hspace{1em} Config.~2   & \textbf{76.667} & \textbf{40.960} & \textbf{12.968} & \textbf{7.673} & 0.018 & \underline{0.011} & \underline{4.511} & \underline{2.918} & 0.9154 & 0.6952 \\
    \rowcolor{maroon!5}   \hspace{1em} Config.~3   & 77.377 & 44.152 & \underline{13.175} & \underline{7.820} & \underline{0.017} & 0.011 & 4.736 & 3.259 & 0.9138 & \underline{0.6950} \\
    \midrule
    \bottomrule\bottomrule
  \end{tabular}%
  }
  \vspace{-0.2em}
\end{table}

\begin{figure}[t]
    \vspace{0em}
    \centering
    \includegraphics[width=\linewidth]{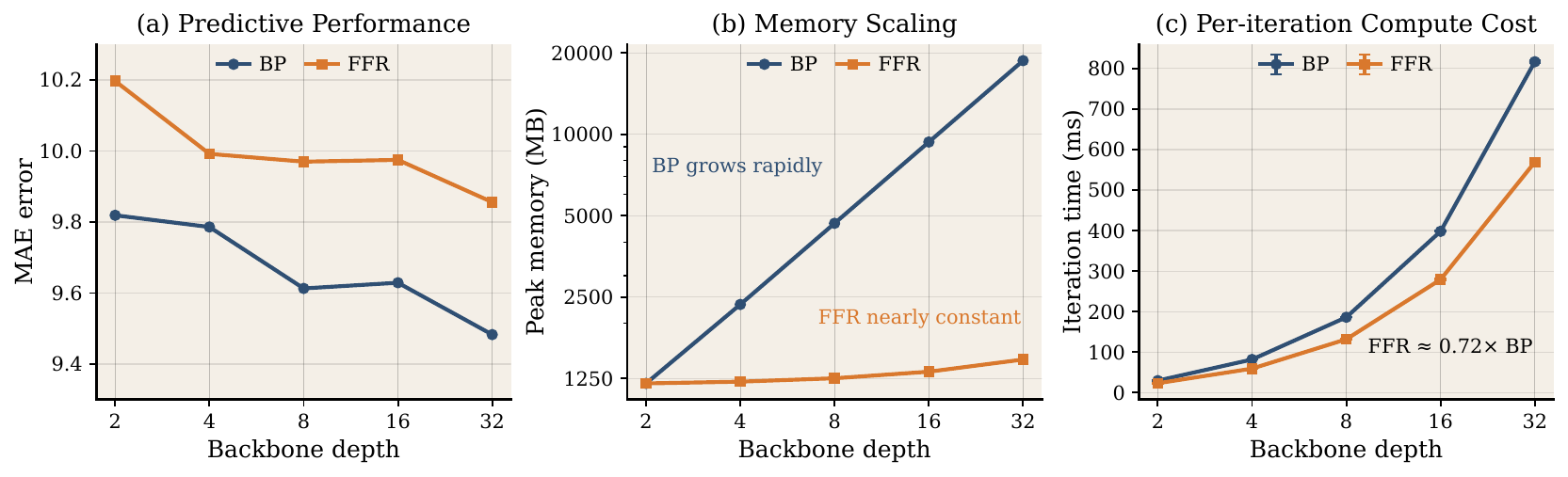}
    \vspace{-1.1em}
    \caption{Scaling of peak training memory and per-iteration training time for FFR vs.\ BP on the KonIQ-10k dataset with a CNN backbone as depth grows. FFR's memory stays flat in depth while BP's grows linearly, and per-iteration compute is reduced relative to BP.}
    \label{fig:scaling}
    \vspace{-1.5em}
\end{figure}

\paragraph{Complexity Analysis.}
Figure~\ref{fig:scaling} compares FFR and BP under increasing model depth on the KonIQ-10k benchmark with the CNN backbone. FFR's per-iteration activation memory stays flat in depth because the layer-local update releases each layer's activations as soon as its local step completes, whereas BP retains every intermediate activation for its backward pass and grows linearly with depth. Training time per iteration is also lower for FFR because the layer-local update skips the input-gradient chain, yielding a $75\%$ asymptotic compute ratio against BP. Moreover, FFR avoids update locking because each layer can update as soon as its local step completes, enabling additional gains from layer-parallel execution. Theoretical and empirical analyses are provided in Appendix~\ref{app:efficiency}, including the single-GPU profiling in Table~\ref{tab:koniq_baseline_profile} and the layer-parallel results in Table~\ref{tab:layer_parallel}.

\begin{figure}[h]
    \vspace{-0.3em}
    \centering
    \includegraphics[width=\linewidth]{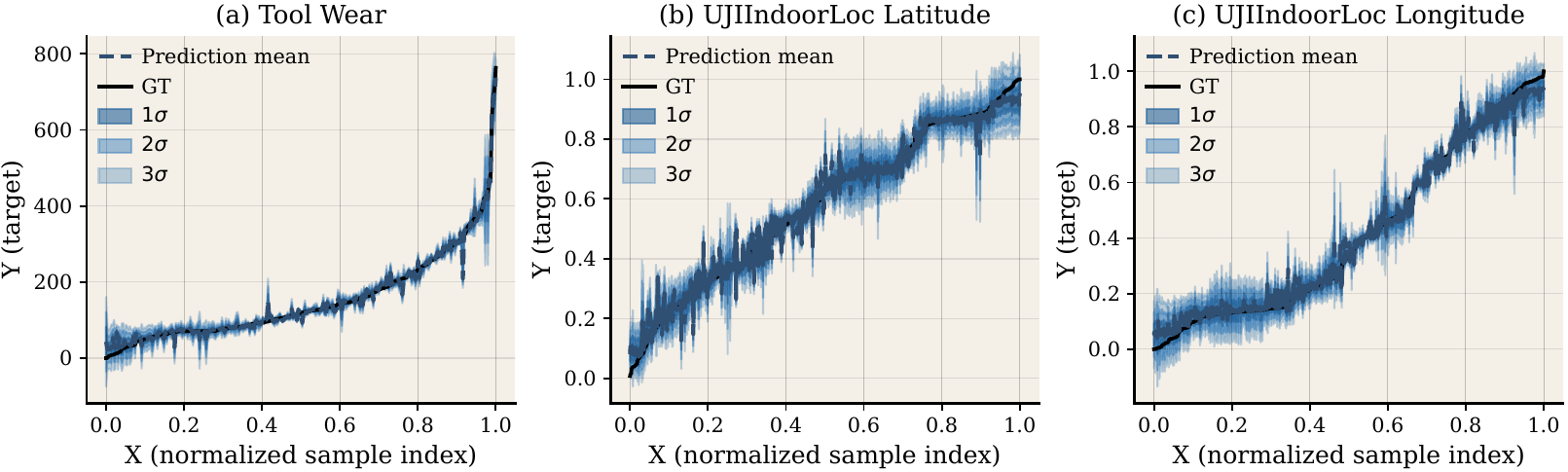}
    \vspace{-1.1em}
    \caption{FFR's predicted mean (dashed) and $1\sigma/2\sigma/3\sigma$ bands against the ground-truth target (solid), with samples sorted by target, where $\sigma$ is FFR's predictive standard deviation.}
    \label{fig:uncertainty}
    \vspace{-1.2em}
\end{figure}

\paragraph{Uncertainty Visualization.}
Figure~\ref{fig:uncertainty} overlays FFR's predicted mean and the $1\sigma/2\sigma/3\sigma$ uncertainty bands on the sorted ground-truth target for Tool Wear and UJIIndoorLoc latitude/longitude, where $\sigma$ is FFR's predictive standard deviation. The mean tracks the target closely, and the bands widen where errors are larger, suggesting that FFR's predictive uncertainty tracks prediction difficulty. Appendix~\ref{app:uncertainty_eval} evaluates the disagreement-based uncertainty score quantitatively against BP-UR with MC Dropout and Snapshot Ensembles.

\section{Conclusion}
\label{sec:conclusion}
We presented FFR, the first framework that effectively enables Forward-Forward learning for real-world regression. To address FF's lack of natural positive/negative pairs and its magnitude-blind goodness, FFR combines three innovations: an ordinal competitive goodness function with distance-aware supervision, a stratified ladder architecture that aggregates multi-scale features into a terminal regression head, and a hierarchical predictor that delivers a robust estimate and uncertainty score in one forward pass. Across six real-world benchmarks spanning smart-home IoT, industrial manufacturing, indoor localization, wearable health, sentiment analysis, and image quality assessment, FFR closely matches BP's accuracy, substantially outperforms all BP-free competitors, keeps peak training memory nearly flat with depth, and delivers per-iteration speedups, showing that carefully designed local learning can rival global optimization at a fraction of the training cost.

{
\small
\vspace{-0.5em}
\section*{Ethics Statement}
\vspace{-0.35em}
This work uses only publicly available benchmark datasets and does not involve the collection of new data from human subjects. BIDMC consists of de-identified physiological recordings released through PhysioNet, and the remaining datasets are standard public benchmarks used under their respective licenses. We do not foresee direct negative societal impacts beyond those generally associated with improving the efficiency of regression models.

\vspace{-0.7em}
\section*{AI Use Statement}
\vspace{-0.35em}
Large language models were used to assist with writing, editing, and formatting the manuscript, as well as with implementing and debugging parts of the experimental code. All methods, experimental designs, and analyses were conceived by the authors, and all LLM-assisted code was reviewed and validated by the authors. The authors remain responsible for the scientific content, experimental results, analyses, and references in this paper.

\vspace{-0.7em}
\section*{Reproducibility Statement}
\vspace{-0.35em}
The method, training procedure, benchmark construction, model architectures, optimization settings, and baseline configurations are described in Sections~\ref{sec:method}--\ref{sec:exp} and Appendices~\ref{app:multitarget}--\ref{app:details}. Additional analyses and efficiency measurements are provided in Appendix~\ref{app:results}. Code will be made publicly available.

\vspace{-0.6em}
\bibliographystyle{iclr2027_conference}
\bibliography{references}
}

\newpage
\appendix
\section{Appendix}
This appendix provides supporting material for the main paper. Appendix~\ref{app:bp_ff} reviews the computational and biological plausibility issues commonly raised against backpropagation and how Forward-Forward addresses them. Appendix~\ref{app:multitarget} details multi-output regression and summarizes the layer-local FFR training procedure. Appendix~\ref{app:uncertainty} provides the theoretical foundation for uncertainty score estimation. Appendix~\ref{app:datasets} describes the benchmark datasets and Appendix~\ref{app:baselines} details every baseline used in our experiments. Appendix~\ref{app:details} reports the full implementation details. Appendix~\ref{app:results} provides additional experimental results, including sensitivity to target range and ordinal discretization (Appendix~\ref{app:sensitivity}), training dynamics and representation collapse (Appendix~\ref{app:results:dynamics}), analysis of FFR's gains over BP-UR (Appendix~\ref{app:ffr_gains}), quantitative uncertainty evaluation (Appendix~\ref{app:uncertainty_eval}), and a per-iteration time/memory comparison between FFR and BP (Appendix~\ref{app:efficiency}).

\subsection{Limitations of BP}
\label{app:bp_ff}

Backpropagation has been criticized for both computational and biological limitations arising from global gradient-based credit assignment \citep{lillicrap2016random,schuman2022opportunities}. Specifically, several aspects of BP create computational bottlenecks or depart from biologically plausible learning:
\begin{enumerate}[label=(\roman*)]
    \item \textbf{Weight transport.} BP's chain rule requires the backward pass to multiply errors by the exact transpose of each forward-pass weight matrix, so an additional feedback pathway must be kept synchronized with the forward weights at all times. Biological circuits have no plausible mechanism for maintaining two physically distinct sets of synapses in lock-step, making this exact weight symmetry, the \emph{weight transport problem}, neurobiologically implausible \citep{lillicrap2016random,nokland2016direct}.
    \item \textbf{Non-local credit assignment.} The gradient with respect to a single synaptic weight in BP aggregates activations and error signals from the entire downstream subnetwork, so each update relies on information that is not available at the synapse itself. In cortex, by contrast, plasticity is driven by signals confined to the pre- and post-synaptic neurons that the synapse physically connects \citep{schuman2022opportunities,yi2023activity}.
    \item \textbf{Frozen network activity.} BP cleanly separates inference from learning into two disjoint phases: errors are routed back through dedicated channels that do not re-enter the forward computation, leaving the previously computed activations frozen until the backward sweep finishes. Biological neurons instead continuously integrate feedback signals from higher areas with their feed-forward inputs, so neural activity continues to evolve during plasticity rather than pausing for a separate backward phase \citep{zhu2025stochastic,yi2023activity}.
    \item \textbf{Update locking.} Because every layer's gradient depends on quantities computed by layers above and below it, no parameter can be updated until the full forward and backward sweep over the current sample has finished, the so-called \emph{update locking problem}. The resulting serial bottleneck blocks layer-parallel and pipelined training and rules out any truly online, sample-by-sample learning regime \citep{nokland2019training,bengio2006greedy,lorberbom2024layer,ma2025ff,sun2025deeperforward}.
\end{enumerate}

The Forward-Forward algorithm \citep{hinton2022forward} structurally addresses all four issues. By replacing the global backward pass with a layer-local goodness objective evaluated on a single forward pass: (i) no feedback pathway with transposed weights is required, removing weight transport; (ii) each layer's update depends only on its own activations and a local target, restoring synaptic locality; (iii) the forward activities themselves drive the weight updates, so the network activity is never frozen by a separate backward phase; (iv) each layer can be updated as soon as its forward activations are available, removing update locking and enabling layer-parallel learning. FFR inherits these properties as a direct extension of FF (Section~\ref{sec:method}).

\subsection{Multi-output regression and training procedure}
\label{app:multitarget}
When extending FFR to a $D$-dimensional target $\mathbf{y}\in\mathbb{R}^D$, only the grouping and ordinal bins of Section~\ref{sec:goodness} are factorized per output dimension; the underlying layer parameters remain shared. Each layer is a single dense map (or convolution, for image inputs) applied identically to the full input for all $D$ targets; only the resulting $n_\ell$ hidden units (or channels) are partitioned into $D$ sets of $K_\ell/D$ competitive groups, one per target dimension. The layer-local objective is the average of the $D$ per-dimension ordinal losses of Eq.~\eqref{eq:ordinal_ce}, so a single shared representation is trained jointly on all targets. At the terminal head (Eq.~\eqref{eq:ladder_reg}), the concatenated goodness vectors from \emph{all} target dimensions are mapped by one linear map to \emph{all} $D$ outputs, so each prediction can draw on the goodness signal of every other target. Cross-target structure is thus captured by the shared backbone and the jointly-trained terminal head. The two multi-target synthetic tasks, MT-A and MT-B (Appendix~\ref{app:datasets}), and the naturally coupled UJIIndoorLoc latitude/longitude targets empirically test this design. As $D$ increases, the backbone cost remains unchanged: the layer parameters are shared, and the per-dimension grouping only re-partitions the existing $n_\ell$ hidden units, dividing the $K_\ell$ groups among targets rather than replicating them, so neither $D$ nor the number of bins adds parameters or computation. The goodness vector fed to the terminal head keeps length $\sum_{\ell} K_\ell$, and only the terminal head grows linearly with $D$. The trade-off lies in per-target ordinal resolution rather than cost, as each output receives $K_\ell/D$ groups, consistent with the bin-resolution sensitivity in Appendix~\ref{app:sensitivity}. Algorithm~\ref{alg:ffr} summarizes the layer-local training procedure and applies to both single-output ($D=1$) and multi-output regression.

\begin{algorithm}[h]
\caption{FFR Training Algorithm}
\label{alg:ffr}
\begin{algorithmic}[1]
\REQUIRE Training set $\mathcal{D}$ with targets $\mathbf{y}\in\mathbb{R}^D$; hidden-layer parameters $\Theta=\{\theta_\ell\}_{\ell=1}^{L-1}$; terminal-head parameters $\theta_{\mathrm{reg}}=(W_{\mathrm{reg}},b_{\mathrm{reg}})$
\FOR{each minibatch $(\mathbf{x},\mathbf{y})\sim\mathcal{D}$}
    \STATE $\mathbf{h}_0 \leftarrow \mathbf{x}$
    \FOR{$\ell=1,\ldots,L-1$}
        \STATE $\mathbf{h}_\ell \leftarrow f_\ell(\mathbf{h}_{\ell-1};\theta_\ell)$
        \STATE Compute grouped goodness and ordinal targets for each output dimension $d=1,\ldots,D$
        \STATE Compute $\mathcal{L}^{\mathrm{ord}}_\ell=\frac{1}{D}\sum_{d=1}^{D}\mathcal{L}^{\mathrm{ord}}_{\ell,d}$
        \STATE Update $\theta_\ell$ using only $\mathcal{L}^{\mathrm{ord}}_\ell$
    \ENDFOR
    \STATE $\mathbf{g}\leftarrow[\mathbf{g}_1;\ldots;\mathbf{g}_{L-1}]$
    \STATE $\hat{\mathbf{y}}_{\mathrm{reg}}\leftarrow W_{\mathrm{reg}}\mathbf{g}+b_{\mathrm{reg}}$
    \STATE Compute $\mathcal{L}^{\mathrm{reg}}=\frac{1}{D}\lVert\hat{\mathbf{y}}_{\mathrm{reg}}-\mathbf{y}\rVert_2^2$
    \STATE Update $\theta_{\mathrm{reg}}$ using only $\mathcal{L}^{\mathrm{reg}}$
\ENDFOR
\STATE \textbf{Locality:} No gradients are backpropagated across layers.
\end{algorithmic}
\end{algorithm}

\subsection{Theoretical foundation of uncertainty estimation}
\label{app:uncertainty}
The uncertainty score in Eq.~\eqref{eq:uncertainty} is motivated by predictor disagreement, a commonly used signal for epistemic uncertainty~\citep{kendall2017uncertainties}. Related approaches estimate uncertainty from variability across stochastic predictions or model ensembles, including MC Dropout~\citep{gal2016dropout}, Deep Ensembles~\citep{lakshminarayanan2017simple}, and snapshot-based ensembles~\citep{huang2017snapshot,jung2023a}. FFR differs in that it exploits the local learning nature of FF: since every layer is trained with its own local objective, each layer is itself a predictor of the target, so multiple predictors arise from a single deterministic network within the same forward pass.

To make this connection precise, let $f_\ell=\mu_\ell$ for $\ell=1,\ldots,L-1$ and $f_L=\hat{y}_{\mathrm{reg}}$. The hierarchical prediction in Eq.~\eqref{eq:ladder_pred} can then be written as
\begin{equation}
    \hat{y}=\sum_{\ell=1}^{L}w_\ell f_\ell,
    \qquad
    w_\ell\geq 0,\qquad
    \sum_{\ell=1}^{L}w_\ell=1.
\end{equation}
Under squared loss, the weighted average error of the individual predictors satisfies
\begin{align}
    \sum_{\ell=1}^{L} w_\ell (y-f_\ell)^2
    &=
    (y-\hat{y})^2
    +
    \sum_{\ell=1}^{L} w_\ell(f_\ell-\hat{y})^2.
\end{align}
Rearranging gives
\begin{equation}
    (y-\hat{y})^2
    =
    \sum_{\ell=1}^{L} w_\ell (y-f_\ell)^2
    -
    U(\mathbf{x}),
\end{equation}
where $U(\mathbf{x})=\sum_{\ell=1}^{L}w_\ell(f_\ell-\hat{y})^2$ is exactly the uncertainty score defined in Eq.~\eqref{eq:uncertainty}. This is the weighted form of the classical ambiguity decomposition~\citep{krogh1994neural}: the ensemble error equals the weighted average error of its members minus their disagreement.

This identity provides a direct interpretation of $U(\mathbf{x})$ as the disagreement among the predictors contributing to the final estimate. In FFR, this disagreement reflects variation among predictors derived from representations learned at different depths and ordinal resolutions. We therefore use $U(\mathbf{x})$ as a single-pass estimate of epistemic uncertainty, rather than as a calibrated predictive variance. Its empirical association with prediction error is evaluated in Appendix~\ref{app:uncertainty_eval}.

\subsection{Dataset descriptions}
\label{app:datasets}

We provide detailed descriptions of the four synthetic function regression tasks and six real-world regression datasets used in our evaluation.

\paragraph{Synthetic function regression.}
Inputs are sampled i.i.d.\ uniformly from $[-1,1]^d$, with $d{=}2$ for \emph{Sin-Cos}, $d{=}5$ for \emph{Exp-Trig-Poly}, and $d{=}7$ for both multi-target tasks.
The two single-target functions are adopted verbatim from \citet{guo2026local} and probe two qualitatively different regimes: \emph{Sin-Cos} is a low-dimensional bounded trigonometric target, while \emph{Exp-Trig-Poly} mixes an exponential-sinusoidal envelope with multiplicative polynomial and trigonometric terms in a higher-dimensional input space:
\begin{align*}
f_{\text{Sin-Cos}}(x_1,x_2) &= \sin(x_1) + \cos(x_2),\\
f_{\text{Exp-Trig-Poly}}(x_1,\dots,x_5) &= e^{x_1}\sin(x_2) + x_3 \cos(x_4) - x_5\, x_1.
\end{align*}
The two multi-target tasks, MT-A and MT-B, take seven inputs $(x_1,\dots,x_7) \in [-1,1]^7$ and are constructed from the same two bases on disjoint coordinate subsets:
\begin{align*}
g_1(x_1,x_2) &= \sin(x_1) + \cos(x_2),\\
g_2(x_3,\dots,x_7) &= e^{x_3}\sin(x_4) + x_5 \cos(x_6) - x_7\, x_3.
\end{align*}
MT-A (two targets) couples the two bases through purely nonlinear combinations, $y_1 = \sin(g_1) + 0.5\,g_2^2$ and $y_2 = g_1\, g_2$, so that no output is a pure base; this stresses whether a model can exploit higher-order correlations between the shared latent factors. MT-B (four targets) introduces private additive contributions per target, $y_1 = g_1 + 0.3\sin(2 x_1)$, $y_2 = 0.7\,g_1 + g_2$, $y_3 = g_2 + 0.5\,x_5^2$, and $y_4 = 0.5\,g_1 + 0.4\cos(x_2) + 0.2\,x_7$, producing a multi-output regression task in which the outputs share substantial latent structure through $g_1$ and $g_2$ but each retains target-specific variation.

\paragraph{Real-world regression datasets.}
Beyond the synthetic benchmarks, we evaluate on six real-world regression datasets spanning smart-home energy management, industrial predictive maintenance, indoor localization, wearable health monitoring, image quality assessment, and text-based sentiment regression.
These datasets span heterogeneous input modalities (physiological waveforms, WiFi RSSI fingerprints, ambient sensor readings, multi-sensor industrial telemetry, natural images, and text token sequences) and differ substantially in sample count and feature dimensionality.
Together they stress whether a BP-free learner can deliver competitive accuracy under the same forward-only, memory-efficient constraints.
The datasets are detailed below.
\begin{itemize}
    \item \textbf{Appliances Energy} \citep{candanedo2017data}: Smart home IoT dataset with 19,735 samples recorded at 10-minute intervals over 4.5 months from a ZigBee wireless sensor network. The 28 input features include temperature and humidity readings from 9 rooms plus weather station data. The regression target is household appliance energy consumption in Wh. This dataset represents the scenario where a home gateway must predict energy usage locally without uploading fine-grained occupancy-revealing sensor data to the cloud.
    \item \textbf{Machine Tool Wear} \citep{de2023tool}: Industrial predictive maintenance dataset from Vicomtech containing 2,054 one-second segments from 13 cutting tools during CNC turning operations. Each segment is described by 153 features extracted (mean, RMS, max, skewness, kurtosis) from a rich multi-sensor array including acoustic emission, two triaxial accelerometers, a three-axis force sensor, spindle and axis motor currents/voltages, a microphone, and CNC-internal signals. Rows containing missing feature values are discarded during preprocessing, leaving 1,483 segments used in our experiments. The regression target is flank wear $V_b$ (mm). This dataset represents the scenario where an industrial edge controller must continuously estimate tool degradation from shop-floor sensor data that cannot leave the factory network due to proprietary constraints.
    \item \textbf{UJIIndoorLoc} \citep{torres2014ujiindoorloc}: WiFi fingerprint-based indoor localization dataset comprising 21,048 samples, each described by 520 WiFi RSSI (Received Signal Strength Indicator) values from detected wireless access points across 3 buildings covering 108,703~m$^2$. The regression targets are longitude and latitude coordinates. This dataset represents the scenario where a home robot must perform localization on its embedded controller without uploading location-tagged WiFi scans that reveal precise user positions.
    \item \textbf{BIDMC PPG and Respiration} \citep{pimentel2017toward}: Wearable health monitoring dataset from PhysioNet containing 53 eight-minute recordings from critically ill patients at Beth Israel Deaconess Medical Center. Each recording includes PPG, impedance respiratory, and ECG signals sampled at 125~Hz, along with 1~Hz numerics (heart rate, SpO2, pulse rate). We construct samples using 8-second windows with 50\% overlap; each window contains 1,000 raw PPG samples and 1,000 raw ECG samples together with age and gender, giving 2,002 input features. This produces 6,032 valid windows, with respiratory rate defined as the mean impedance-respiration annotation over the same 8-second interval. This dataset represents the scenario where a wearable pulse oximeter must estimate respiratory rate on-device, as continuous physiological waveforms are highly sensitive and should not leave the device.
    \item \textbf{KonIQ-10k} \citep{hosu2020koniq}: A large-scale blind image quality assessment benchmark of 10,073 images sampled from a public photographic database to span a broad range of authentic distortions (compression, noise, blur, exposure, etc.). Each image is annotated with a Mean Opinion Score (MOS) on a 1--5 scale aggregated from over 120 crowd-sourced ratings per image, which is rescaled to the $1$–$100$ range. The regression target is the per-image MOS. We use an $80/20$ train/held-out split, resize images to $64{\times}48$, and normalize pixels with ImageNet statistics. KonIQ-10k provides a natural test of FFR's transfer to a high-dimensional vision regression task with a CNN backbone.
    \item \textbf{CMU-MOSEI} \citep{zadeh2018multimodal}: A multimodal sentiment regression benchmark with targets in $[-3,3]$; we use the text modality from the utterance transcripts. We follow the official training, validation, and test splits with 16,274, 1,861, and 4,653 utterances, respectively. Text is lowercased and tokenized using the regular expression \texttt{[a-z0-9']+}; the vocabulary is constructed from the training split only and retains the 4,000 most frequent entries including \texttt{<pad>} and \texttt{<unk>}. Each utterance is truncated or padded to 40 tokens and mapped through a 128-dimensional embedding. No feature scaling or missing-value filtering is applied. CMU-MOSEI evaluates whether FFR transfers from feed-forward backbones to a recurrent GRU backbone.
\end{itemize}

\subsection{Baseline descriptions}
\label{app:baselines}

We provide detailed descriptions of all baseline methods:

\begin{itemize}
    \item \textbf{BP-UR (BP upper reference)}: Standard end-to-end backpropagation with the same FFR architecture (group normalization and multi-scale feature aggregation included) trained with a global MSE loss, serving as the performance upper reference.
    \item \textbf{BP-EX (BP with extra loss)}: BP-UR augmented with the per-layer ordinal losses of FFR (Eq.~\eqref{eq:ordinal_ce}), applied on top of the global MSE objective.
    \item \textbf{FF-MSE}: An intuitive FF-based baseline that uses grouped neurons and multi-scale channel aggregation, with each layer supervised greedily by a direct per-group MSE loss between its activations and the regression target (see Section~\ref{sec:goodness}).
    \item \textbf{FF-CLF}: An intuitive FF-based baseline that uses grouped neurons and multi-scale channel aggregation, with each layer supervised by a one-hot grouped-classification loss obtained by discretizing the regression target into bins.
    \item \textbf{FF-CAR (classification-as-regression)} \citep{padmani2025function}: An existing FF method that extends FF to regression via cosine-similarity goodness and in-/out-tol probe sampling.
    \item \textbf{Trifecta} \citep{dooms2023trifecta}: An existing FF method introducing three simple techniques (normalization, new goodness, and contrastive loss) that stabilize and deepen FF networks.
    \item \textbf{FF-Zero} \citep{guo2026local}: An existing FF-style method that replaces the FF goodness with directional-derivative estimation in favor of physical-hardware feasibility.
    \item \textbf{PEPITA} \citep{dellaferrera2022error}: A non-FF BP-free method that, instead of a backward pass, performs a second forward pass on the input perturbed by a fixed random projection of the output error and updates hidden layers from the modulated activations.
    \item \textbf{$\text{F}^3$ (Feed-Forward with delayed Feedback)} \citep{flugel2024f3}: A non-FF BP-free method that approximates gradients via fixed random feedback paths and uses delayed sample-wise error from the previous epoch as a scaling factor, addressing both the weight-transport and update-locking problems while preserving forward-only training.
\end{itemize}

\subsection{Implementation details}
\label{app:details}
\paragraph{MLP backbone for tabular data.}
All methods share a 4-layer fully connected backbone: three FF-trainable hidden blocks of width $256$, followed by a terminal regression head.
Each hidden block consists of a linear map, a GELU activation, and group-wise normalization.
The input dimensionality is dataset-specific (2 for Sin-Cos, 5 for Exp-Trig-Poly, 7 for MT-A and MT-B, 28 for Appliances Energy, 153 for Machine Tool Wear, 520 for UJIIndoorLoc, and 2002 for BIDMC).
For FFR, the three hidden layers use group counts $K_\ell = (16, 32, 64)$, the terminal head is a linear map from the concatenated grouped goodness of hidden layers to the target.

\paragraph{CNN backbone for image regression.}
For the KonIQ-10k benchmark we replace the FC backbone with an 8-layer convolutional network. Each FF-trainable conv layer is a $3{\times}3$ convolution with ReLU and group-wise normalization; all channel widths are $256$. Per-group goodness $g_{\ell,k}$ is the spatially averaged squared activation of the channels assigned to group $k$, so the ordinal competitive objective of Eq.~\eqref{eq:ordinal_ce} applies with no other change. All eight conv layers are FF-trained with group counts $K_\ell = (16, 32, 64)$ for the first three layers and $K_\ell = 64$ for layers $4$--$8$; the per-layer goodness vectors are concatenated across all eight layers and fed to a linear regression head predicting the image MOS. Input images are resized to $64{\times}48$ and normalized with ImageNet statistics. All baselines use the same backbone and training budget, except PEPITA and FF-CAR. PEPITA fails to converge once the CNN backbone is deepened beyond a few layers, so we report it under a shallower CNN that still trains stably. FF-CAR is not formulated for convolutional architectures, and its in-tol/out-tol probe sweep makes both training and inference prohibitively expensive on CNN inputs at this resolution; we therefore report it under its native (non-CNN) configuration where feasible and as ``--'' otherwise.

\paragraph{GRU backbone for sequential regression.}
For CMU-MOSEI, we use three GRU blocks of width $256$ with group counts $K_\ell=(16,32,64)$. The first block receives 128-dimensional token embeddings, while subsequent blocks operate on the preceding hidden sequence. Group normalization is applied within each block, and each group's score is obtained by averaging the squared normalized GRU outputs over time and the channels assigned to that group. The three grouped-score vectors are concatenated and fed to a linear terminal regression head. In FFR, each GRU block is optimized using its own local ordinal objective, while the terminal head is trained separately on detached grouped-score features so that no gradients are backpropagated between recurrent blocks.

\paragraph{FFR-specific hyperparameters.}
We use distance-aware soft ordinal labels with bandwidth $\sigma_\ell=\Delta_\ell$ (Eq.~\eqref{eq:soft_label}), the group schedule $K_\ell$ that progressively sharpens the ordinal targets with depth and base exponent $d_0 = 4$ (Section~\ref{sec:ladder}), and the hierarchical prediction weights of Eq.~\eqref{eq:ladder_pred} set uniformly. We additionally cap $K_\ell$ at half of the layer's hidden width, so that each competitive group still contains at least two neurons or kernels; for multi-target datasets, $K_\ell$ is divided by the number of targets so that the per-target group count remains comparable to the single-target setting.

\paragraph{Optimization.}
Unless explicitly noted, all methods are optimized with Adam (default PyTorch $\beta_1 = 0.9$, $\beta_2 = 0.999$) for 500 epochs.
The synthetic and four tabular datasets (Appliances Energy, Tool Wear, UJIIndoorLoc, BIDMC) use learning rate $10^{-3}$ and batch size $512$ on Appliances Energy and UJIIndoorLoc; Tool Wear keeps the same learning rate but switches to a smaller batch size of $64$ and trains for 1000 epochs to compensate for its limited data size; BIDMC uses learning rate $3\times 10^{-4}$ and batch size $512$.
KonIQ-10k uses learning rate $10^{-3}$, batch size $64$, and 200 epochs on its 8-layer CNN backbone. CMU-MOSEI uses learning rate $10^{-3}$, batch size $64$, and up to 100 epochs.
Each FF-style layer is trained using only its local objective, with no gradients backpropagated between layers. The terminal regression head is trained by MSE on the concatenated goodness vectors without backpropagating into the hidden layers.
No learning-rate scheduling, weight decay, or gradient clipping is applied.

\paragraph{Data splits and preprocessing.}
UJIIndoorLoc and CMU-MOSEI follow their official dataset splits. For the remaining datasets, Appliances Energy, Machine Tool Wear, and KonIQ-10k use an $80/20$ train/held-out split, and BIDMC is split by subject. All compared methods use the same dataset splits for a fair comparison. For the four real-world tabular datasets, each input feature is min--max scaled to $[-1,1]$ using the minimum and maximum computed from the training split; held-out inputs are transformed using the same training statistics and clipped to $[-1,1]$. Regression targets are rescaled to $[0,1]$ during training and mapped back to the original scale at evaluation time.
The ordinal bins used by FFR and all classification-style baselines span the training-split target range.

\paragraph{Baseline configurations.}
BP (upper reference) and BP (extra loss) share the exact FFR architecture (including group normalization and ladder aggregation) and training budget.
For BP (extra loss), the auxiliary per-layer ordinal loss uses the same group schedule $K_\ell = (16, 32, 64)$ as FFR and is weighted by $0.3$, with $0.7$ on the global MSE.
All classification-style BP-free baselines (FF-CLF, PEPITA, Trifecta) discretize the target into $64$ uniform bins across the training-split target range and recover a scalar prediction as $\sum_k p_k m_k$, where $p_k$ are the predicted bin logits and $m_k$ is the midpoint of bin $k$.
FF-MSE uses the same backbone with MSE loss as the goodness function at every hidden layer.
FF-CAR follows the configuration of \citet{padmani2025function} with cosine-similarity goodness. All experiments are run in PyTorch on a single NVIDIA RTX 3090 GPU (24~GB VRAM).

\subsection{Additional experimental results}
\label{app:results}
\subsubsection{Sensitivity to target range and ordinal discretization}
\label{app:sensitivity}

We evaluate FFR's sensitivity to the target range and ordinal discretization. The target ranges vary substantially across datasets, from $[0,1]$ for UJIIndoorLoc to $[10,1080]$ Wh for Appliances, while all synthetic targets lie in $[-1,1]$. Unless otherwise stated, we use the same bin schedule $K_\ell=(16,32,64)$ across datasets without dataset-specific tuning.

\paragraph{Target range and extrapolation.}
To evaluate sensitivity to the training target range, we restrict training to samples within the central $80\%$ of the target range and evaluate on samples outside this range. Table~\ref{tab:range_shift} reports the original and shifted-test RMSE. All methods degrade under target-range extrapolation; even BP-UR loses one to two orders of magnitude, reflecting the general difficulty of predicting beyond the observed target range. FFR nevertheless remains the strongest BP-free method across all four tasks and is markedly more robust to the shift than the other BP-free methods, with the largest margins on the multi-target benchmarks. We attribute this robustness to the stratified ladder architecture, which supplies predictors at multiple ordinal scales, and to the hierarchical prediction.

\begin{table}[t]
\centering
\small
\caption{Sensitivity to target-range extrapolation. Values are original / out-of-range test RMSE.}
\label{tab:range_shift}
\resizebox{0.72\linewidth}{!}{%
\begin{tabular}{lcccc}
\toprule
Method & Sin-Cos & Exp-Trig-Poly & MT-A & MT-B \\
\midrule
BP-UR & 0.005 / 0.203 & 0.007 / 0.564 & 0.008 / 0.578 & 0.011 / 0.150 \\
BP-EX & 0.013 / 0.344 & 0.014 / 0.548 & 0.012 / 0.606 & 0.019 / 0.145 \\
F$^3$ & 0.096 / 0.384 & 0.078 / 0.751 & 0.067 / 0.858 & 0.091 / 0.288 \\
PEPITA & 0.490 / 0.820 & 0.106 / 0.656 & 0.112 / 1.027 & 0.206 / 0.528 \\
Trifecta & 0.026 / 0.366 & 0.035 / 0.682 & 0.048 / 0.768 & 0.095 / 0.287 \\
FF-Zero & 0.137 / 0.465 & 0.140 / 0.929 & 0.151 / 1.131 & 0.231 / 0.643 \\
FF-MSE & 0.196 / 0.529 & 0.152 / 0.803 & 0.172 / 1.002 & 0.253 / 0.634 \\
FF-CLF & 0.010 / 0.378 & 0.027 / 0.659 & 0.108 / 1.010 & 0.365 / 0.833 \\
FFR (ours) & 0.005 / 0.305 & 0.009 / 0.581 & 0.015 / 0.681 & 0.021 / 0.223 \\
\bottomrule
\end{tabular}%
}
\end{table}

\paragraph{Bin resolution.}
We next vary the layer-wise bin schedule while keeping the hidden width fixed. Table~\ref{tab:bin_resolution} shows that FFR is insensitive to the bin schedule as long as it is not the finest one: RMSE stays flat from $\{4,8,16\}$ to the default $\{16,32,64\}$ and degrades only at $\{32,64,128\}$, where FFR still outperforms the best BP-free baseline in Table~\ref{tab:synthetic} and Table~\ref{tab:realworld}. Because the hidden width is fixed, the finest schedule splits the same neurons into more and smaller competitive groups, so each group's goodness is averaged over fewer units. Keeping at least four units per group is thus a practical rule of thumb, and pushing to the finest bin resolution is unnecessary.

\begin{table}[t]
\centering
\small
\caption{Sensitivity to the layer-wise bin schedule. Values are test RMSE.}
\label{tab:bin_resolution}
\begin{tabular}{lcccc}
\toprule
$K_\ell$ schedule & MT-A & MT-B & Appliances & Tool Wear \\
\midrule
$\{4,8,16\}$ & 0.016 & 0.015 & 76.513 & 13.621 \\
$\{8,16,32\}$ & 0.017 & 0.019 & 76.201 & 12.840 \\
$\{16,32,64\}$ (default) & 0.015 & 0.021 & 76.777 & 13.547 \\
$\{32,64,128\}$ & 0.024 & 0.026 & 82.036 & 16.901 \\
\bottomrule
\end{tabular}
\end{table}

\begin{figure}[!b]
\vspace{-0.5em}
    \centering
        \begin{minipage}[t]{0.49\linewidth}
        \centering
        \includegraphics[width=\linewidth]{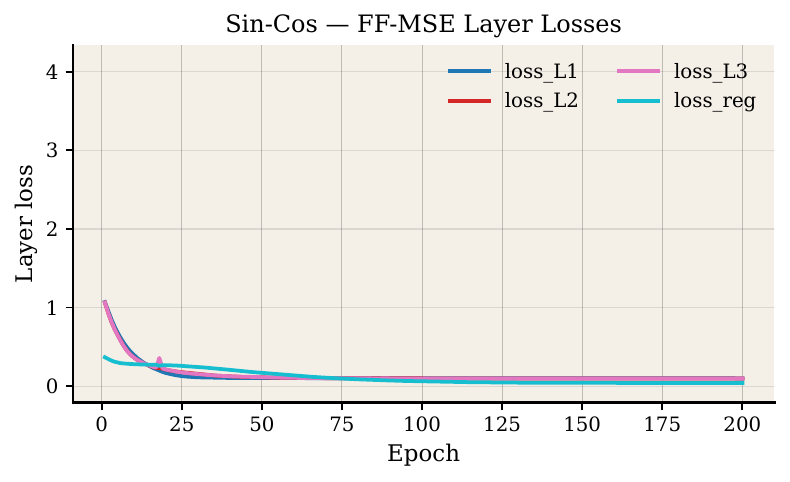}
    \end{minipage}
    \hfill
    \begin{minipage}[t]{0.49\linewidth}
        \centering
        \includegraphics[width=\linewidth]{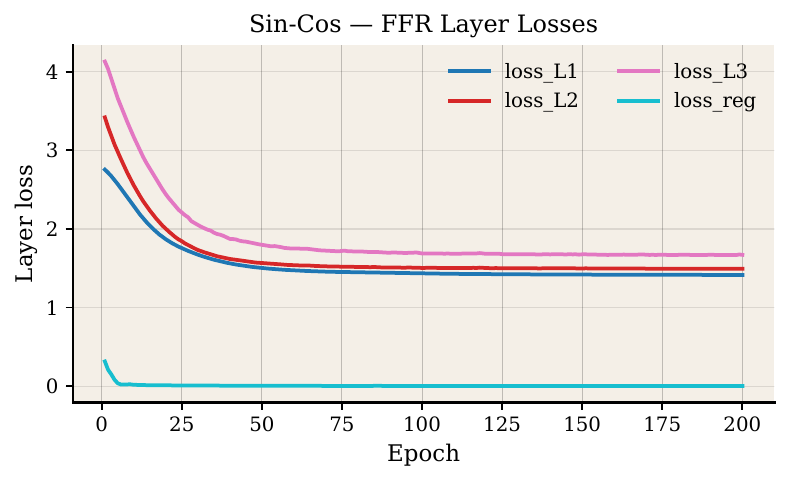}
    \end{minipage}
    \\[0.4em]
    \begin{minipage}[t]{0.49\linewidth}
        \centering
        \includegraphics[width=\linewidth]{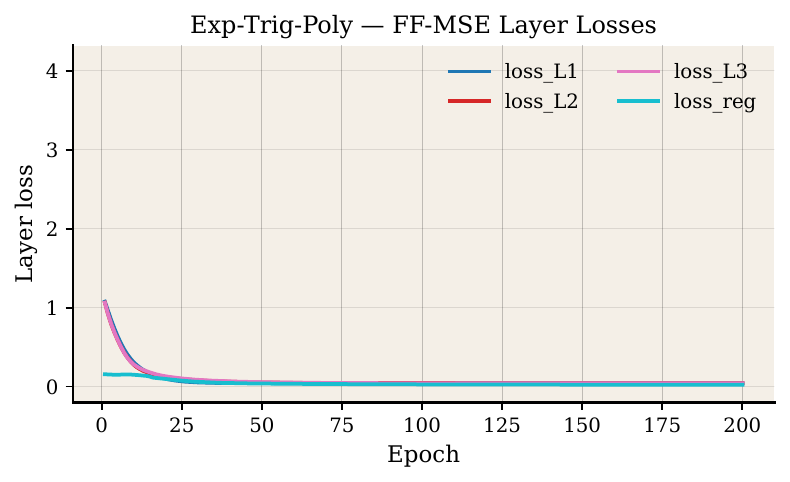}
    \end{minipage}
    \hfill
    \begin{minipage}[t]{0.49\linewidth}
        \centering
        \includegraphics[width=\linewidth]{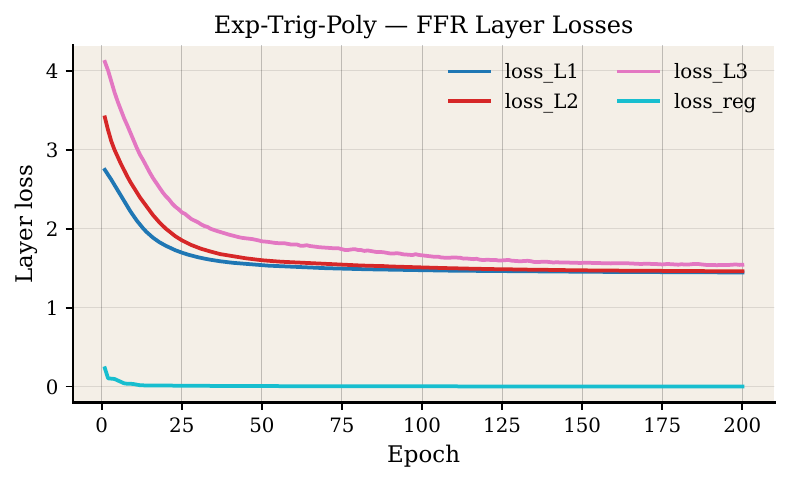}
    \end{minipage}

    \caption{Per-layer training loss curves on the two synthetic tasks.}
    \label{fig:layer_losses}
\vspace{-0.5em}
\end{figure}

\paragraph{Non-uniform bins.}
Finally, we vary the bin widths according to $\Delta_k\propto k^\alpha$, linearly rescaled to span the full target range, where $\alpha=0$ recovers uniform bins and a larger $\alpha$ makes the bins grow wider toward one end of the range; the bandwidth in Eq.~\eqref{eq:soft_label} is set to the width of the corresponding bin. Table~\ref{tab:nonuniform_bins} shows that the effect of non-uniform binning is dataset-dependent. On the real-world Appliances and Tool Wear datasets, non-uniform bins remain robust and can even improve performance: with $\alpha=0.5$, the RMSE drops from $76.777$ to $76.070$ on Appliances and from $13.547$ to $12.432$ on Tool Wear. In contrast, the synthetic multi-target tasks degrade monotonically as $\alpha$ increases. A plausible explanation is the target distribution. Appliances and Tool Wear have the most skewed targets among the four datasets (mean skewness $3.39$ and $2.13$), so non-uniform bin widths better match their target density, whereas MT-A and MT-B are much less skewed ($1.11$ and $0.04$), where uniform bins remain more effective.

\begin{table}[t]
\centering
\small
\caption{Effect of non-uniform bin widths. Values are test RMSE.}
\label{tab:nonuniform_bins}
\begin{tabular}{lccc}
\toprule
Dataset & Uniform & $\alpha=0.5$ & $\alpha=1.0$ \\
\midrule
MT-A & 0.015 & 0.022 & 0.037 \\
MT-B & 0.021 & 0.033 & 0.054 \\
Appliances & 76.777 & 76.070 & 77.121 \\
Tool Wear & 13.547 & 12.432 & 13.615 \\
\bottomrule
\end{tabular}
\end{table}

\begin{table}[t]
\centering
\small
\caption{Mean cross-layer linear CKA ($\downarrow$ indicates more differentiated representations). The random-Gaussian null baseline is $0.114$.}
\label{tab:cka}
\begin{tabular}{lcccc}
\toprule
Method & Sin-Cos & Exp-Trig-Poly & MT-A & MT-B \\
\midrule
FF-MSE & 0.979 & 0.689 & 0.520 & 0.924 \\
FFR & 0.709 & 0.518 & 0.377 & 0.555 \\
BP-UR & 0.704 & 0.546 & 0.466 & 0.487 \\
\bottomrule
\end{tabular}
\vspace{-0.8em}
\end{table}

\subsubsection{Training dynamics and representation collapse}
\label{app:results:dynamics}
To empirically verify the representation collapse discussed in Section~\ref{sec:goodness}, we collect per-layer diagnostics over the full training trajectory on the two synthetic function regression tasks (Sin-Cos and Exp-Trig-Poly).
First, we plot the per-layer training-loss trajectories of FFR and FF-MSE on the two synthetic tasks (Figure~\ref{fig:layer_losses}). FF-MSE's hidden layers all collapse onto nearly the same trajectory and converge rapidly together, indicating that every layer regresses the target in essentially the same way, a signature of representation collapse under greedy regression that simultaneously hints at overfitting at every depth. FFR's layers, by contrast, follow visibly distinct trajectories at different paces: shallow layers converge quickly to coarse ordinal predictions while deeper layers continue to refine, evidencing cooperative coarse-to-fine learning across depth rather than redundant copies of the same predictor.

\begin{figure}[!b]
    \vspace{-1.0em}
    \centering
    \includegraphics[width=\linewidth]{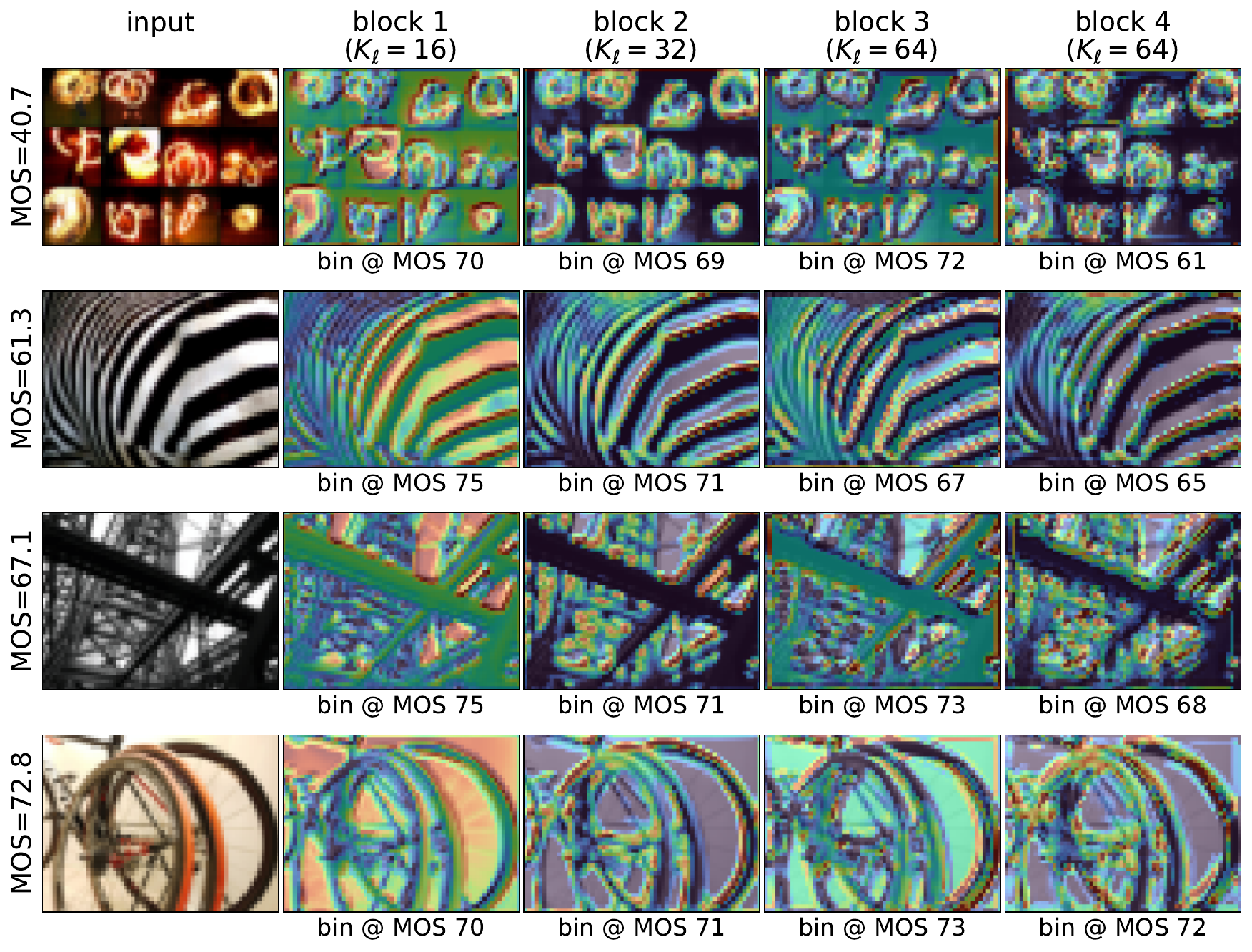}
    \caption{Spatial activation maps of FFR's winning competitive group across the stratified ladder on KonIQ-10k. Rows show four randomly selected test images with ground-truth MOS values $40.7$, $61.3$, $67.1$, and $72.8$ from top to bottom. Columns show the input and four FF-trained blocks with $K_\ell=(16,32,64,64)$. For each block, channels in the highest-goodness group are averaged over the feature map, upsampled to image resolution, and overlaid on a grayscale copy of the input. The label below each map gives the MOS at the center of the winning bin. Each panel is independently scaled by its 2nd--98th activation percentiles.}
    \label{fig:koniq_spatial}
    \vspace{-1.0em}
\end{figure}

\paragraph{Cross-layer representation similarity.}
We further quantify representation collapse using linear centered kernel alignment (CKA)~\citep{kornblith2019similarity}, where larger values indicate more similar representations across layers. Table~\ref{tab:cka} reports the mean cross-layer CKA on 2,000 samples; random Gaussian features give a null baseline of $0.114$. FF-MSE consistently exhibits the highest cross-layer similarity, whereas FFR is substantially lower and follows the same overall trend as BP-UR. This supports the training-dynamics observation that FFR learns progressively differentiated representations rather than redundant copies of the same regressor.

\paragraph{Layer-wise spatial and ordinal structure.}
To visualize how the differentiated representations manifest in the CNN backbone, we randomly select four KonIQ-10k test images. Figure~\ref{fig:koniq_spatial} shows the winning-group activation for the four images with ground-truth MOS $40.7$, $61.3$, $67.1$, and $72.8$. The activation patterns differ visibly across depth, with different blocks emphasizing different contours and spatial regions of the same image. The winning ordinal bins also vary across blocks ($70/69/72/61$, $75/71/67/65$, $75/71/73/68$, and $70/71/73/72$ for the four rows, respectively), showing that the layer-wise predictors are not identical copies of one another.

\begin{figure}[t]
    \centering
    \includegraphics[width=\linewidth]{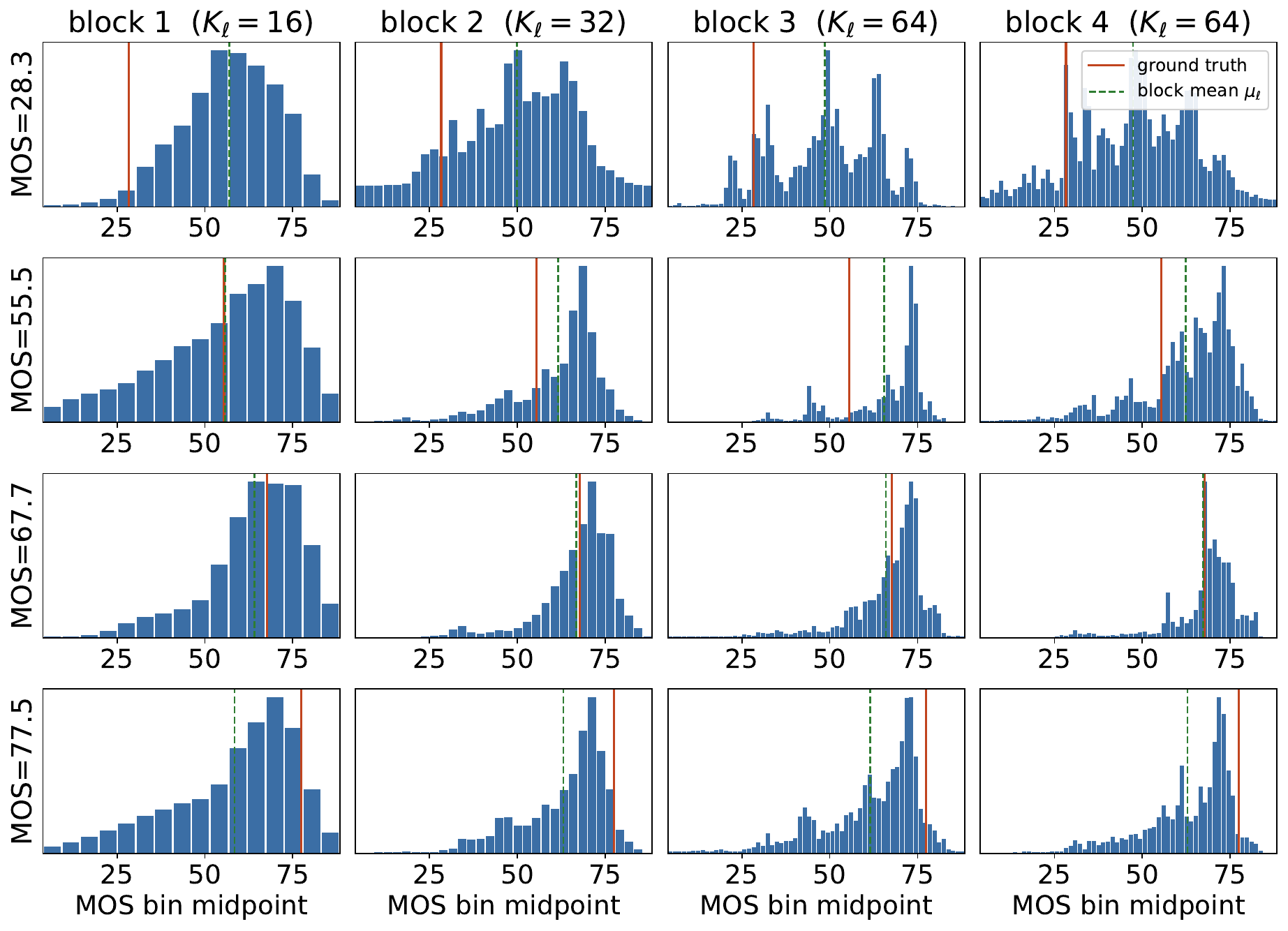}
    \caption{Per-block ordinal bin distributions of FFR on KonIQ-10k. Columns show the four blocks with $K_\ell=(16,32,64,64)$. Bars show $p_{\ell,k}=\mathrm{softmax}(g_{\ell,k})$ at bin midpoints $m_{\ell,k}$, with bar width equal to the corresponding bin width. The solid line marks the ground-truth MOS and the dashed line the block prediction $\mu_\ell=\sum_k p_{\ell,k}m_{\ell,k}$.}
    \label{fig:koniq_bins}
\end{figure}

Figure~\ref{fig:koniq_bins} provides the complementary prediction-space view through the per-block ordinal distributions $p_{\ell,k}$. As the bin resolution increases with depth, the distributions do not simply become finer copies of the shallow distribution: deeper blocks develop distinct, often multi-modal structures and their continuous predictions $\mu_\ell$ shift across depth. The distributions are comparatively sharp and aligned with the target for the accurately predicted third image, while the lowest-MOS example remains broader and more dispersed. Together with the lower cross-layer CKA of FFR, these visualizations support the interpretation that different blocks learn non-redundant features and predictors that can provide complementary signals to hierarchical aggregation.

\paragraph{Convergence speed.}
We further compare how quickly the methods approach their own converged solutions. For each method, let $L_{\min}$ denote the minimum value of its training criterion, and record the first epoch at which it falls below $1.5L_{\min}$, $1.2L_{\min}$, and $1.1L_{\min}$ (Table~\ref{tab:convergence_speed}). FFR uses the training RMSE of the final hierarchical prediction, while the baselines use the training loss of their final predictor. Because each method is measured relative to its own $L_{\min}$, the comparison concerns convergence speed rather than absolute loss values.

\begin{table}[t]
\centering
\small
\caption{Epochs required to reach $1.5L_{\min}$ / $1.2L_{\min}$ / $1.1L_{\min}$ of each method's own converged training criterion ($\downarrow$). For KonIQ-10k, wall-clock times for the same thresholds are also reported.}
\label{tab:convergence_speed}
\resizebox{\linewidth}{!}{
\begin{tabular}{lcccccc}
\toprule
Method & Appliances & Tool Wear & UJIIndoorLoc & BIDMC & KonIQ-10k & KonIQ-10k time \\
\midrule
BP-UR  & 407 / 413 / 443 & 502 / 515 / 528 & 48 / 155 / 160 & 256 / 408 / 424 & 68 / 138 / 157 & 291.7 / 592.0 / 673.5 s \\
FFR    & 172 / 309 / 375 & 649 / 767 / 856 & 46 / 91 / 143 & 240 / 276 / 305 & 39 / 99 / 139 & 122.1 / 309.9 / 435.1 s \\
FF-MSE & 124 / 248 / 319 & 896 / 961 / 969 & 14 / 67 / 109 & 393 / 449 / 476 & 34 / 87 / 133 & 114.2 / 292.3 / 446.9 s \\
FF-CLF & 28 / 121 / 208 & 542 / 552 / 563 & 17 / 93 / 163 & 199 / 319 / 408 & 41 / 56 / 111 & 135.3 / 184.8 / 366.3 s \\
\bottomrule
\end{tabular}
}
\end{table}

FFR reaches all three thresholds earlier than BP-UR on Appliances, BIDMC, and KonIQ-10k, and reaches the two tighter thresholds earlier on UJIIndoorLoc. Tool Wear is the main exception, where FFR requires more epochs. On KonIQ-10k, using the measured per-epoch costs of $4.29$ s for BP-UR, $3.13$ s for FFR, $3.36$ s for FF-MSE, and $3.30$ s for FF-CLF, FFR reaches the three thresholds in approximately $42\%$, $52\%$, and $65\%$ of BP-UR's wall-clock time. Overall, FFR does not systematically trade its lower per-iteration cost for more training epochs. The results also reveal a characteristic difference between local and global learning. The FF-based methods reduce their training criterion rapidly in the early stage, often reaching $1.5L_{\min}$ within relatively few epochs (e.g., $28$--$172$ on Appliances versus $407$ for BP-UR), while the subsequent refinement toward $1.1L_{\min}$ is slower. Since every layer receives its own target-supervised signal from the start, local learning makes fast initial progress but may require additional optimization for fine-grained convergence. BP-UR, whose layers are coordinated only through the global error signal, reaches the loose threshold later but closes the gap between thresholds more quickly (e.g., $407/413/443$ on Appliances).

\begin{figure}[thbp]
\vspace{-0.5em}
    \centering
    \begin{minipage}[t]{\linewidth}
        \centering
        \includegraphics[width=\linewidth]{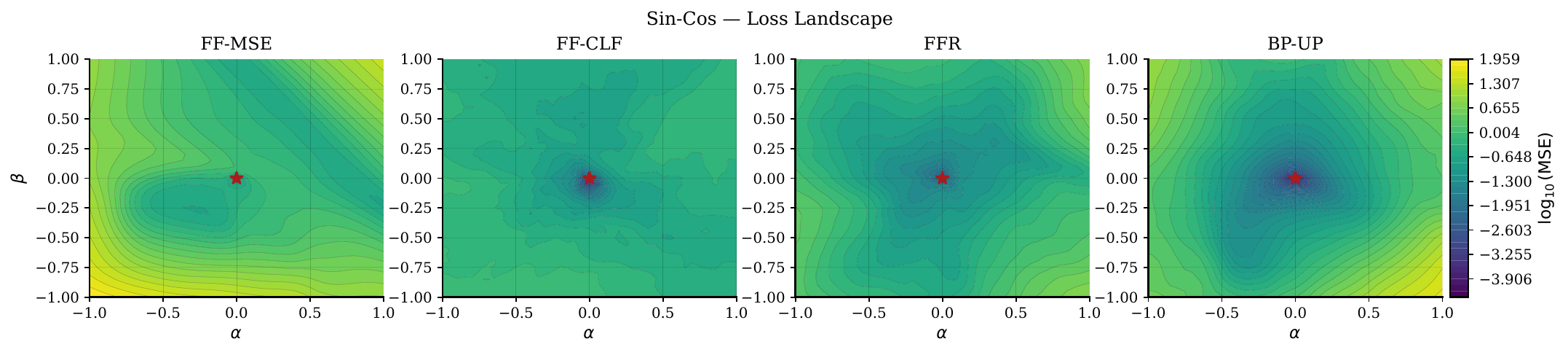}
    \end{minipage}
    \\[0.2em]
    \begin{minipage}[t]{\linewidth}
        \centering
        \includegraphics[width=\linewidth]{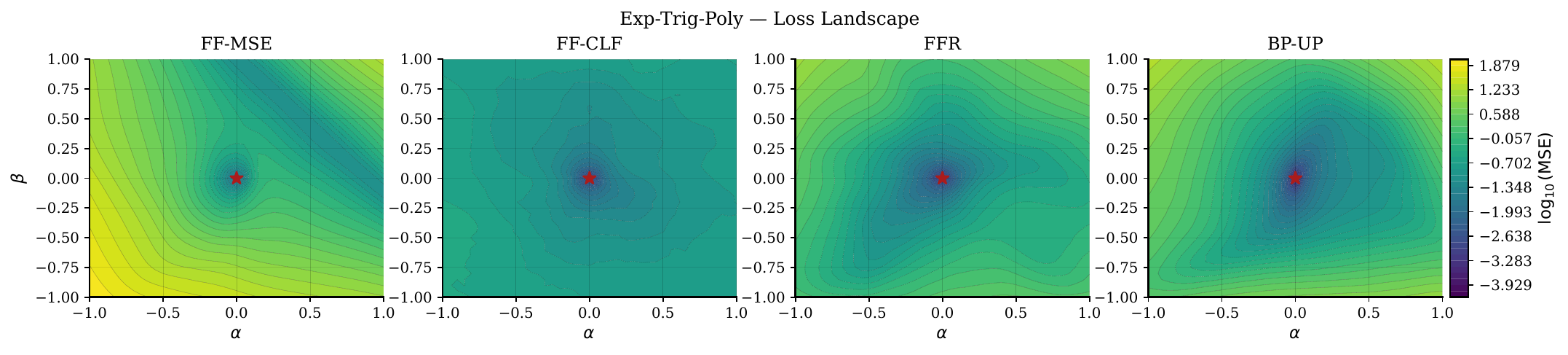}
    \end{minipage}
    \caption{Loss landscape around the converged solution along two random directions, each rescaled to the norm of the corresponding parameter tensors~\citep{NEURIPS2018_a41b3bb3}, on the two synthetic tasks. FFR converges to a flat basin comparable to that of BP, whereas FF-MSE collapses to a sharp minimum.}
    \label{fig:loss_landscape}
\vspace{-1em}
\end{figure}

We also report the loss-landscape visualizations at convergence on the two synthetic tasks (Figure~\ref{fig:loss_landscape}). FF-MSE collapses to a sharp minimum, mirroring the per-layer overfitting observed in its training curves; FF-CLF settles in a comparatively narrow basin; BP and FFR both reach flat basins, and FFR's landscape closely mirrors BP's. This geometric similarity indicates that FFR's per-layer cooperative ordinal objectives serve as an effective surrogate for BP's global optimization target, while inheriting the flat-minimum geometry that is commonly associated with better generalization.

\subsubsection{Analysis of FFR's gains over BP-UR}
\label{app:ffr_gains}

FFR occasionally outperforms BP-UR despite optimizing its hidden layers without global backpropagation. We examine this behavior through the interaction between hierarchical ensembling and cross-layer diversity. Under squared loss, the ambiguity decomposition in Appendix~\ref{app:uncertainty} gives $E=\bar{E}-A$, where $\bar{E}$ is the weighted average error of the individual predictors and $A$ is their weighted disagreement. Thus, hierarchical aggregation can improve upon its constituent predictors when they provide complementary predictions.

\paragraph{Empirical ensemble gain.}
We first verify the ambiguity decomposition numerically on the four synthetic benchmarks. Table~\ref{tab:ambiguity_decomposition} reports the weighted average member error, ambiguity, and final hierarchical prediction error. The residual of $E-(\bar{E}-A)$ is below $10^{-11}$ in all cases, while the ambiguity term offsets approximately $43\%$--$68\%$ of the weighted average member error.

\begin{table}[t]
\centering
\small
\caption{Empirical ambiguity decomposition for FFR. Values are test-set mean squared errors.}
\label{tab:ambiguity_decomposition}
\begin{tabular}{lcccc}
\toprule
& Sin-Cos & Exp-Trig-Poly & MT-A & MT-B \\
\midrule
$\bar{E}$: average member error & $1.01{\times}10^{-4}$ & $1.10{\times}10^{-4}$ & $9.26{\times}10^{-4}$ & $4.14{\times}10^{-3}$ \\
$A$: ambiguity & $6.88{\times}10^{-5}$ & $4.76{\times}10^{-5}$ & $6.21{\times}10^{-4}$ & $2.65{\times}10^{-3}$ \\
$E$: hierarchical error & $3.20{\times}10^{-5}$ & $6.19{\times}10^{-5}$ & $3.05{\times}10^{-4}$ & $1.49{\times}10^{-3}$ \\
\bottomrule
\end{tabular}
\end{table}

\paragraph{Aggregation alone is insufficient.}
The component ablation in Table~\ref{tab:ab_component} shows that removing hierarchical prediction consistently degrades FFR on the real-world benchmarks. To test whether the same aggregation mechanism alone explains the gain, we additionally apply hierarchical prediction to BP-EX, the closest BP-based counterpart with per-layer grouped predictions. As shown in Table~\ref{tab:bpex_hp}, hierarchical prediction improves BP-EX on the two single-target synthetic tasks but degrades performance on both multi-target tasks. This comparison indicates that hierarchical aggregation is useful to FFR but is not, by itself, sufficient to reproduce its gains. The quality and complementarity of the constituent predictors also matter.

\begin{table}[t]
\centering
\small
\caption{Effect of applying hierarchical prediction (HP) to BP-EX. Values are test RMSE ($\downarrow$).}
\label{tab:bpex_hp}
\begin{tabular}{lcccc}
\toprule
Method & Sin-Cos & Exp-Trig-Poly & MT-A & MT-B \\
\midrule
BP-EX & 0.013 & 0.014 & \textbf{0.012} & \textbf{0.019} \\
BP-EX + HP & 0.009 & 0.010 & 0.019 & 0.039 \\
FFR (ours) & \textbf{0.005} & \textbf{0.009} & 0.015 & 0.021 \\
\bottomrule
\end{tabular}
\end{table}

\paragraph{Cross-layer diversity.}
The representation analysis in Appendix~\ref{app:results:dynamics} provides further evidence for this interpretation. Table~\ref{tab:cka} shows substantially stronger cross-layer similarity for FF-MSE than for FFR; for example, on Sin-Cos the average CKA is $0.979$ for FF-MSE, compared with $0.709$ for FFR and $0.704$ for BP-UR. In contrast to this representation collapse, FFR maintains greater differentiation across depth. Its coarse-to-fine ordinal objectives therefore provide layer-wise predictors with more complementary information for hierarchical aggregation.

Taken together, these results suggest that FFR's occasional gains over BP-UR arise from the interaction between coarse-to-fine local supervision and hierarchical aggregation. The gain is not explained by aggregation alone: FFR first learns differentiated predictors across depth, and the hierarchical prediction can then exploit their complementarity.

\subsubsection{Quantitative uncertainty evaluation}
\label{app:uncertainty_eval}

\paragraph{Association with prediction error.}
We next measure the Spearman correlation between each uncertainty estimate and the absolute prediction error. We compare FFR with MC Dropout~\citep{gal2016dropout} using $T=10$ stochastic forward passes and Snapshot Ensembles~\citep{huang2017snapshot,jung2023a} using $M=5$ predictors. Higher correlation indicates that uncertain predictions are more consistently ranked among the larger errors.

\begin{table}[t]
\centering
\small
\caption{Spearman correlation between uncertainty estimates and absolute prediction error ($\uparrow$). The cost column reports inference overhead for the MLP backbone relative to one deterministic forward pass.}
\label{tab:uncertainty_correlation}
\begin{tabular}{lccccc}
\toprule
Method & Cost & Sin-Cos & Exp-Trig-Poly & MT-A & MT-B \\
\midrule
FFR, Eq.~\eqref{eq:uncertainty} & $1\times$ & \textbf{0.498} & \textbf{0.286} & \textbf{0.742} & \textbf{0.834} \\
BP-UR + MC Dropout ($T=10$) & $5.5\times$ & 0.247 & 0.268 & 0.438 & 0.377 \\
BP-UR + Snapshot Ensembles ($M=5$) & $5\times$ & 0.230 & 0.254 & 0.379 & 0.344 \\
\bottomrule
\end{tabular}
\end{table}

Table~\ref{tab:uncertainty_correlation} reports the results on the four synthetic benchmarks. Averaged over these four tasks, the Spearman correlation is $0.590$ for FFR, compared with $0.333$ for MC Dropout and $0.302$ for Snapshot Ensembles, and FFR achieves the highest correlation on every task. FFR obtains $U(\mathbf{x})$ from the same forward pass used for prediction. MC Dropout incurs approximately $5.5\times$ inference overhead on the MLP backbone. These results evaluate the ranking induced by the uncertainty score rather than its absolute calibration; post-hoc calibration on a held-out set can be applied when calibrated intervals are required.

\subsubsection{Training efficiency: computation and memory}
\label{app:efficiency}

We analyze the per-iteration computational and memory complexity of Forward-Forward (FF) and backpropagation (BP) under the assumption that both methods require a comparable number of optimization steps to reach a target performance. Therefore, we focus on the cost of a single iteration. The measurements in Figure~\ref{fig:background}(c), Figure~\ref{fig:scaling}, and Table~\ref{tab:koniq_baseline_profile} use sequential execution on a single GPU. The theoretical ratio derived below likewise assumes no layer parallelism.

\paragraph{Notation.}
Consider a fully connected network with $L$ layers. For layer $\ell$, let
\begin{equation}
x_{\ell-1} \in \mathbb{R}^{H}, \quad
W_\ell \in \mathbb{R}^{H \times H}, \quad
z_\ell = W_\ell x_{\ell-1}, \quad
x_\ell = \sigma(z_\ell),
\end{equation}
where $\sigma(\cdot)$ denotes the ReLU activation function. Let $\delta x_\ell \in \mathbb{R}^{H}$ denote the gradient signal at layer $\ell$. We measure computational cost in terms of dominant operations, where matrix--vector products and outer products scale as $O(H^2)$, while element-wise operations scale as $O(H)$.

\paragraph{Per-layer computational complexity.}
For FF, each layer performs a linear forward pass, activation, local gradient computation, weight gradient construction, and parameter update:
\begin{equation}
z_\ell = W_\ell x_{\ell-1}, \quad
x_\ell = \mathrm{ReLU}(z_\ell), \quad
\delta z_\ell = \delta x_\ell \odot \mathbf{1}_{z_\ell > 0},
\end{equation}
\begin{equation}
g_{W_\ell} = \delta z_\ell x_{\ell-1}^{\top}, \quad
W_\ell \leftarrow W_\ell - \eta g_{W_\ell}.
\end{equation}
The corresponding computational cost is
\begin{equation}
T_{\mathrm{FF}}^{\ell} = 3O(H^2) + 2O(H).
\end{equation}

For BP, the same operations are performed, together with gradient propagation to the previous layer:
\begin{equation}
\delta x_{\ell-1} = W_\ell^\top \delta z_\ell,
\end{equation}
which incurs an additional $O(H^2)$ cost. Therefore, the per-layer complexity of BP is
\begin{equation}
T_{\mathrm{BP}}^{\ell} = 4O(H^2) + 2O(H).
\end{equation}

\paragraph{Total computational complexity.}
For a network with $L$ layers, the total per-iteration cost becomes
\begin{equation}
T_{\mathrm{FF}} = L\left(3O(H^2) + 2O(H)\right),
\end{equation}
\begin{equation}
T_{\mathrm{BP}} = L\left(4O(H^2) + 2O(H)\right).
\end{equation}
When $H$ is sufficiently large, the quadratic terms dominate, yielding
\begin{equation}
\frac{T_{\mathrm{FF}}}{T_{\mathrm{BP}}}
=
\frac{3H^2 + O(H)}{4H^2 + O(H)}
\;\xrightarrow[H \to \infty]{}\;
\frac{3}{4}.
\end{equation}
Thus, FF reduces the dominant computation by a constant factor compared to BP.

\paragraph{Peak memory complexity.}
We next analyze the peak activation memory required during training. In BP, intermediate activations $\{x_\ell\}_{\ell=1}^{L}$ must be stored during the forward pass in order to compute gradients in the backward pass. This results in a peak memory cost of
\begin{equation}
\mathcal{M}_{\mathrm{BP}} = O(LH).
\end{equation}
In contrast, FF performs layer-wise local updates and does not require storing activations across layers. At any time, only the current layer's activations need to be retained, leading to
\begin{equation}
\mathcal{M}_{\mathrm{FF}} = O(H).
\end{equation}
Therefore, FF reduces the peak activation memory by a factor of $O(L)$ compared to BP.

\paragraph{Discussion.}
Both FF and BP share the same asymptotic computational complexity $O(LH^2)$, as their cost is dominated by dense linear operations. However, FF removes the need for cross-layer gradient propagation, reducing the number of matrix-scale operations per layer and yielding a constant-factor computational advantage. More importantly, FF significantly reduces peak memory usage by eliminating the need to store intermediate activations across layers, which is particularly beneficial for deep networks.

\begin{table}[thbp]
\vspace{-1em}
\centering
\caption{KonIQ-10k baseline profiling results (8-layer CNN setting).}
\label{tab:koniq_baseline_profile}
\begin{tabular}{lcc}
\toprule
Method & Time (ms/iter) & Peak Memory (MB) \\
\midrule
BP-UR    & 186.10 & 4687.48 \\
BP-EX    & 193.44 & 4754.79 \\
Trifecta & 402.11 & 3973.83 \\
FF-Zero  & 467.42 & 1255.04 \\
PEPITA   & 155.79 & 1371.46 \\
F3       & 258.87 & 3584.75 \\
FF-MSE   & 130.76 & 1251.88 \\
FF-CLF   & 131.00 & 1252.67 \\
FFR      & 131.81 & 1253.27\\
\bottomrule
\end{tabular}
\vspace{-1em}
\end{table}

\begin{table}[tb]
\centering
\small
\caption{Layer-parallel training efficiency on KonIQ-10k. Relative cost is normalized by single-GPU BP-UR within this experiment.}
\label{tab:layer_parallel}
\begin{tabular}{lccc}
\toprule
Method & GPUs & Time (ms/iter) & Relative cost \\
\midrule
BP-UR & 1 & 32.8 & 100\% \\
FFR, sequential & 1 & 25.1 & 76\% \\
FFR, pipelined & 1 & 22.9 & 70\% \\
\midrule
BP-UR, data parallel & 2 & 17.8 & 54\% \\
FFR, layer parallel & 2 & 14.2 & 43\% \\
BP-UR, data parallel & 4 & 15.2 & 46\% \\
FFR, layer parallel & 4 & 9.4 & 29\% \\
\bottomrule
\end{tabular}
\end{table}

\paragraph{Empirical comparison.} Table~\ref{tab:koniq_baseline_profile} reports per-iteration wall-clock time and peak training memory of every BP-free baseline together with the two BP references on the 8-layer KonIQ-10k CNN setting. BP-UR and BP-EX cluster around $185$--$193$\,ms/iteration and $\approx 4.6$--$4.8$\,GB peak memory, dominated by the activations stored for the backward pass. The naive FF baselines FF-MSE and FF-CLF are the fastest and most memory-efficient ($\approx 132$\,ms/iteration and $\approx 1.24$\,GB, roughly $70\%$ of BP's time and $27\%$ of its memory) since they avoid the backward pass entirely; PEPITA sits between these two regimes. FF-Zero pays a substantial time penalty for repeated forward passes per update of its training scheme; Trifecta is both slow (same repeated forward passes) and high peak memory, because it embeds the target bin class into each input image through an extra FC embedding layer whose embedded feature scales with the image resolution. $\text{F}^3$ uses higher peak memory due to its delayed-feedback buffers. FFR inherits the forward-only profile of FF-MSE and FF-CLF (Figure~\ref{fig:scaling}), achieving comparable per-iteration time and memory to the most efficient BP-free baselines while attaining BP-comparable accuracy in Table~\ref{tab:realworld}.

\paragraph{Layer-parallel execution.}
FFR's layer-local updates also remove the update-locking constraint of BP, allowing computation and parameter updates from different layers to overlap. Following the layer-parallel execution principle explored in prior FF work~\citep{sun2025deeperforward}, we evaluate this potential separately on KonIQ-10k using an 8-layer CNN with 256 channels and batch size 64 on NVIDIA GH200 GPUs. These measurements use a different hardware and execution setup from the single-GPU RTX 3090 profiling above, and are therefore reported as a separate comparison.

Relative to sequential single-GPU FFR, layer-parallel execution provides $1.77\times$ and $2.67\times$ speedups with two and four GPUs, respectively. Under the same GPU count, FFR is $1.25\times$ faster than data-parallel BP-UR with two GPUs and $1.62\times$ faster with four GPUs. These results show that removing update locking provides an additional source of training efficiency beyond the single-GPU computational advantage analyzed above.

\end{document}